\documentclass[letterpaper]{article} 
\usepackage{aaai23}  
\usepackage{times}  
\usepackage{helvet}  
\usepackage{courier}  
\usepackage[hyphens]{url}  
\usepackage{graphicx} 
\usepackage{natbib}  
\usepackage{caption} 
\usepackage{algorithm}
\usepackage{algorithmic}

\usepackage{newfloat}
\usepackage{listings}
\DeclareCaptionStyle{ruled}{labelfont=normalfont,labelsep=colon,strut=off} 
\floatstyle{ruled}
\newfloat{listing}{tb}{lst}{}
\floatname{listing}{Listing}
\usepackage{amsmath,amsfonts,amssymb,amsthm}
\usepackage{booktabs}
\usepackage{multirow}
\usepackage{makecell}
\usepackage[caption=false]{subfig}

\def\log{\text{log}}

\def\sign{\text{sign}}

\def\Ber{\text{Bernoulli}}

\def\GC{\text{GC}}

\def\KL{\text{KL}}

\def\AC{\text{AC}}
\def\MUL{\text{MUL}}
\def\WIP{\text{W-IP}}

\def\A{\mbox{\bf{A}}}
\def\D{\mbox{\bf{D}}}
\def\E{\mathcal{E}}
\def\G{\mathcal{G}}
\def\g{\text{\bf{g}}}
\def\HH{\mbox{\bf{H}}}
\def\I{\mbox{\bf{I}}}

\def\NN{\mathbb{N}}

\def\R{\mathbb{R}}

\def\W{\mbox{\bf{W}}}
\def\X{\mbox{\bf{X}}}
\def\Z{\mbox{\bf{Z}}}

\def\w{\boldsymbol{w}}
\def\x{\boldsymbol{x}}

\def\z{\boldsymbol{z}}
\def\PP{\text{P}}
\def\T{\text{T}}

\def\ppi{\boldsymbol{\pi}}
\def\eeta{\boldsymbol{\eta}}
\newsavebox\CBox

\title{Spiking Variational Graph Auto-Encoders for\\
	Efficient Graph Representation Learning}
\author {
    Hanxuan Yang,\textsuperscript{\rm 1,\rm 2}
    Ruike Zhang, \textsuperscript{\rm 1,\rm 2}
    Qingchao Kong \textsuperscript{\rm 1,\rm 2}
    Wenji Mao \textsuperscript{\rm 1,\rm 2}
}
\affiliations {
    \textsuperscript{\rm 1} School of Artificial Intelligence, University of Chinese Academy of Sciences\\
    \textsuperscript{\rm 2} SKL-MCCS, Institute of Automation, Chinese Academy of Sciences\\
    yanghanxuan2020@ia.ac.cn, zhangruike2020@ia.ac.cn, qingchao.kong@ia.ac.cn, wenji.mao@ia.ac.cn
}
\nocopyright

\begin{document}

\maketitle

\begin{abstract}
Graph representation learning is a fundamental research issue and benefits a wide range of applications on graph-structured data. Conventional artificial neural network-based methods such as graph neural networks (GNNs) and variational graph auto-encoders (VGAEs) have achieved promising results in learning on graphs, but they suffer from extremely high energy consumption during training and inference stages. Inspired by the bio-fidelity and energy-efficiency of spiking neural networks (SNNs), recent methods attempt to adapt GNNs to the SNN framework by substituting spiking neurons for the activation functions. However, existing SNN-based GNN methods cannot be applied to the more general multi-node representation learning problem represented by link prediction. Moreover, these methods did not fully exploit the bio-fidelity of SNNs, as they still require costly multiply-accumulate (MAC) operations, which severely harm the energy efficiency. To address the above issues and improve energy efficiency, in this paper, we propose an SNN-based deep generative method, namely the Spiking Variational Graph Auto-Encoders (S-VGAE) for efficient graph representation learning. To deal with the multi-node problem, we propose a probabilistic decoder that generates binary latent variables as spiking node representations and reconstructs graphs via the weighted inner product. To avoid the MAC operations for energy efficiency, we further decouple the propagation and transformation layers of conventional GNN aggregators. We conduct link prediction experiments on multiple benchmark graph datasets, and the results demonstrate that our model consumes significantly lower energy with the performances superior or comparable to other ANN- and SNN-based methods for graph representation learning.
\end{abstract}

\section{Introduction}

Learning representations on graph-structured data is a fundamental issue in a wide range of domains, such as social network analysis, molecular generation and recommender systems. Due to the powerful representation learning ability, many deep learning-based methods for graphs have been proposed, such as graph neural networks (GNNs) \cite{kipf2017semi, hamilton2017inductive, velivckovic2018graph, liu2020towards} and variational graph auto-encoders (VGAEs) \cite{kipf2016variational, grover2019graphite, sarkar2020graph}. Although some of these methods have achieved promising performances on downstream applications, the success of these artificial neural network (ANN) based methods is at the great cost of energy consumption during both the training and inference stages.

Inspired by the spatio-temporal dynamics mechanism of biological brains, the spiking neural networks (SNNs) offer a promising solution to greatly reduce energy consumption, hence SNNs are viewed as the third generation of neural networks \cite{maass1997networks}. Unlike the conventional ANNs in which the neurons communicate by continuous floating-point numbers, SNNs take discrete spiking signals as the inputs and outputs of each neuron and thus have tremendous potential for energy-saving on neuromorphic hardwares \cite{stockl2021optimized}. For example, on the TrueNorth digital chip \cite{merolla2014million}, SNNs can achieve 100,000$\times$ reduction for energy and 100$\times$ reduction for time consumption compared with ANNs \cite{cassidy2014real}. Recently, many SNN-based methods have been proposed for downstream applications on Euclidean data, such as image classification \cite{stockl2021optimized, skatchkovsky2021learning}, object detection \cite{kim2020spiking} and image generation \cite{kamata2022fully}. However, few SNN-based methods have modeled the graph-structured data or been designed for graph representation learning.

To apply SNNs to graph-structured data, recent work \cite{xu2021exploiting, zhu2022spiking} attempts to adapt conventional GNNs to the SNN framework by substituting spiking neurons such as the leaky integrate-and-fire (LIF) model \cite{stein1967frequency} for the activation functions. These SNN-based GNN methods have achieved obvious efficiency improvement in node classification task compared with the previous ANN-based methods. However, there still remain two critical challenges for graph representation learning within the SNN framework. (1) \textbf{Multi-node representation learning}. Existing SNN-based methods rely on GCNs which can only handle single-node tasks such as node classification, whereas the multi-node representation learning problem is largely ignored, in which \textit{link prediction} is regarded as a most important task in practice \cite{zhang2021labeling, jo2021edge}. (2) \textbf{Energy efficiency}. To learn the representations of graph nodes, existing SNN-based methods employ conventional GNN aggregators with floating-point coefficients to perform neighbor aggregation, which require costly multiply-accumulate (MAC) operations that are inefficient to implement on the neuromorphic hardware.

\begin{figure}[ht]
	\begin{center}
		\centerline{\includegraphics[width=\columnwidth]{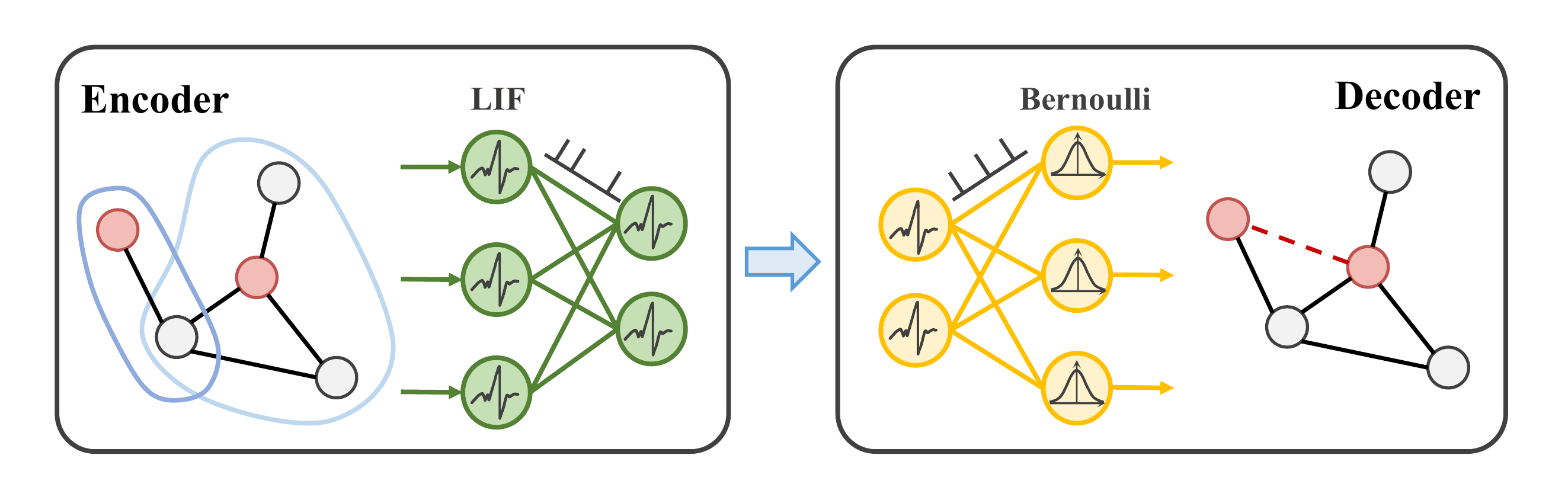}}\par
		\caption{Schematic view of our proposed S-VGAE.}
		\label{intro}
	\end{center}
\end{figure}

To tackle the above research challenges, in this paper, we propose the SNN-based Spiking Variational Graph Auto-encoders (S-VGAE) for efficient graph representation learning. In contrast to the existing SNN-based GNN methods, our proposed method adopts the deep generative framework for graphs based on VGAEs, and adapts it to the SNN framework with a spiking GNN encoder and a probabilistic spiking decoder. To address the representative link prediction task for the multi-node problem, the probabilistic spiking decoder first generates Bernoulli latent variables as the spiking node representations for graph reconstruction. Then, a novel \textit{weighted inner product} (W-IP) readout layer for SNNs is proposed to measure the similarity between binary spiking node representations (see Fig.~\ref{intro}). To avoid MAC floating-point operations (FLOPs) and improve energy efficiency, the spiking GNN encoder decouples conventional GNN aggregators into the stacked SNN propagation and transformation layers. The \textit{propagation} layer aggregates neighbor information based on the graph topology and the \textit{transformation} layer performs linear transformations via the trainable synaptic weights. Both layers employ spiking neurons to emit spiking signals, which can transform the expensive MAC operations to energy-efficient accumulate (AC) operations.

The main contributions of this paper are as follows:
\begin{itemize}
	\item We propose the first SNN-based deep generative method, namely the Spiking Variational Graph Auto-encoders (S-VGAE), for energy-efficient graph representation learning.
	\item For link prediction in the multi-node problem, we design the probabilistic spiking decoder with the weighted inner product (W-IP) layer to enable the similarity computation between spiking node representations.
	\item For energy efficiency, we design the spiking GNN encoder to transform the expensive MAC operations to low-cost AC operations by decoupling conventional GNN aggregators into propagation and transformation layers.
	\item Experiments for link prediction on multiple benchmark graph datasets show that our proposed method achieves significantly lower energy consumption with performances better or comparable to the ANN- and SNN-based methods.
\end{itemize}

\section{Related Work}

We briefly review the literature related to graph representation learning methods and SNNs.

\subsection{Graph Neural Networks}

With the development of deep learning, the end-to-end GNNs have achieved considerable success in learning on graph-structured data \cite{kipf2017semi, hamilton2017inductive, velivckovic2018graph, wu2019simplifying, liu2020towards, zhang2021labeling, jo2021edge}. These methods employ an aggregator to learn the neighbor information of graph nodes. For instance, GCN \cite{kipf2017semi} employs the normalized Laplacian to average node features over all one-hop neighbors. The graph attention network (GAT) \cite{velivckovic2018graph} leverages the parameterized attention coefficients to endow node neighbors with different weights. Despite the good graph representation learning ability, these methods are typically designed for some specific tasks and lack generalizability for multi-node representation learning problems.

\subsection{Deep Generative Methods for Graphs}

With the success of GNNs, there has been emerging interest in learning representations on graph data using the deep generative methods as well \cite{kipf2016variational, wang2018graphgan, grover2019graphite, sarkar2020graph, zheng2020distribution}. These methods typically include an inferential or discriminative model parameterized by deep leaning-based methods (such as GNNs), and an unsupervised generative model for graph reconstruction, which can be used for general multi-node tasks, including link prediction \cite{kipf2016variational, sarkar2020graph, zheng2020distribution}, triplet classification \cite{dai2020generative}, and motif identification \cite{lin2021generative}. For example, based on the variational auto-encoder (VAE) \cite{kingma2014auto} framework, the variational graph auto-encoder (VGAE) \cite{kipf2016variational} employ GCNs to learn latent variables as node representations and an inner product (IP) layer to reconstruct graphs for link prediction with similar distributions to the inputs. Compared with GNNs, these deep generative methods enjoy better robustness and generalizability \cite{zheng2020distribution} and have shown promising results on multiple graph representation learning problems, including single-node tasks such as node classification \cite{grover2019graphite}, and multi-node tasks represented by link prediction \cite{sarkar2020graph}.

\subsection{Spiking Neural Networks}

SNNs are brain-inspired neural networks that mimic the structure and information-propagating mechanism of biological brains. The neurons in SNNs take event-driven spiking signals as inputs and outputs, which are much more bio-plausible and energy-saving. In addition, SNNs are intrinsically dynamic with abundant temporal information conveyed by spike timing. However, the non-differentiable spiking signals also bring difficulties for training deep SNNs by disabling backpropagation that significantly impair the performances of SNNs. Therefore, early methods are mainly ``converted'' SNNs, which attempt to convert pre-trained ANN frameworks to SNN versions by replacing the continuous activation functions with spiking neurons \cite{rueckauer2017conversion, sengupta2019going, kim2020spiking, stockl2021optimized, bu2022optimal}. However, most of these converted SNNs require long latency with a large number of time steps, which severely increases the amount of computation. In recent years, there are more and more ``directly trained'' SNNs \cite{wu2018spatio, wu2019direct, zheng2021going, xu2021exploiting, fang2021deep, kamata2022fully, zhu2022spiking}, which employ the surrogate gradient as smooth approximations of the non-differentiable signals to implement backpropagation and have shown superior performances over the converted SNNs on Euclidean data \cite{zheng2021going, fang2021deep}.

Unfortunately, there are still few SNN-based methods that can learn representations on graph-structured data. Existing methods attempt to adapt GNNs to the SNN framework for node classification, whereas the multi-node tasks have been neglected. Moreover, these methods still require floating-point MAC operations to perform neighbor aggregation, which can largely harm the energy efficiency of SNNs. For instance, GraphSNN \cite{xu2021exploiting} employs GNN aggregators with floating-point coefficients to learn representations of single nodes. SpikingGCN \cite{zhu2022spiking} leverages the simple graph convolution (SGC) \cite{wu2019simplifying} framework and thus needs to compute the power of normalized graph Laplacian during encoding, which leads to a large amount of MACs. Unlike these methods, in this paper, we propose an SNN-based deep generative method S-VGAE for graph representation learning. Based on the VGAE framework, our method employs a probabilistic decoder to address the representative multi-node task, i.e., link prediction. Furthermore, to avoid MACs and improve energy efficiency, we decouple the GNN aggregators into the propagation and transformation layers for the encoder. To the best of our knowledge, this is the first work to establish a deep generative model for graph representation learning within the SNN framework.

\section{Method}

In this section, we propose the SNN-based deep generative S-VGAE method for graph representative learning. The input data include an undirected graph $\G$ with $N$ nodes and a node feature matrix $\X=(\x_1,\dots,\x_N)'\in\R^{N\times C_0}$, where $C_0$ is the input dimension. Our objective is to learn the spiking node representations that can reconstruct the input graph as well as possible. 

\subsection{Spiking Variational Graph Auto-Encoder}

\begin{figure}[ht]
	\begin{center}
		\centerline{\includegraphics[width=\columnwidth]{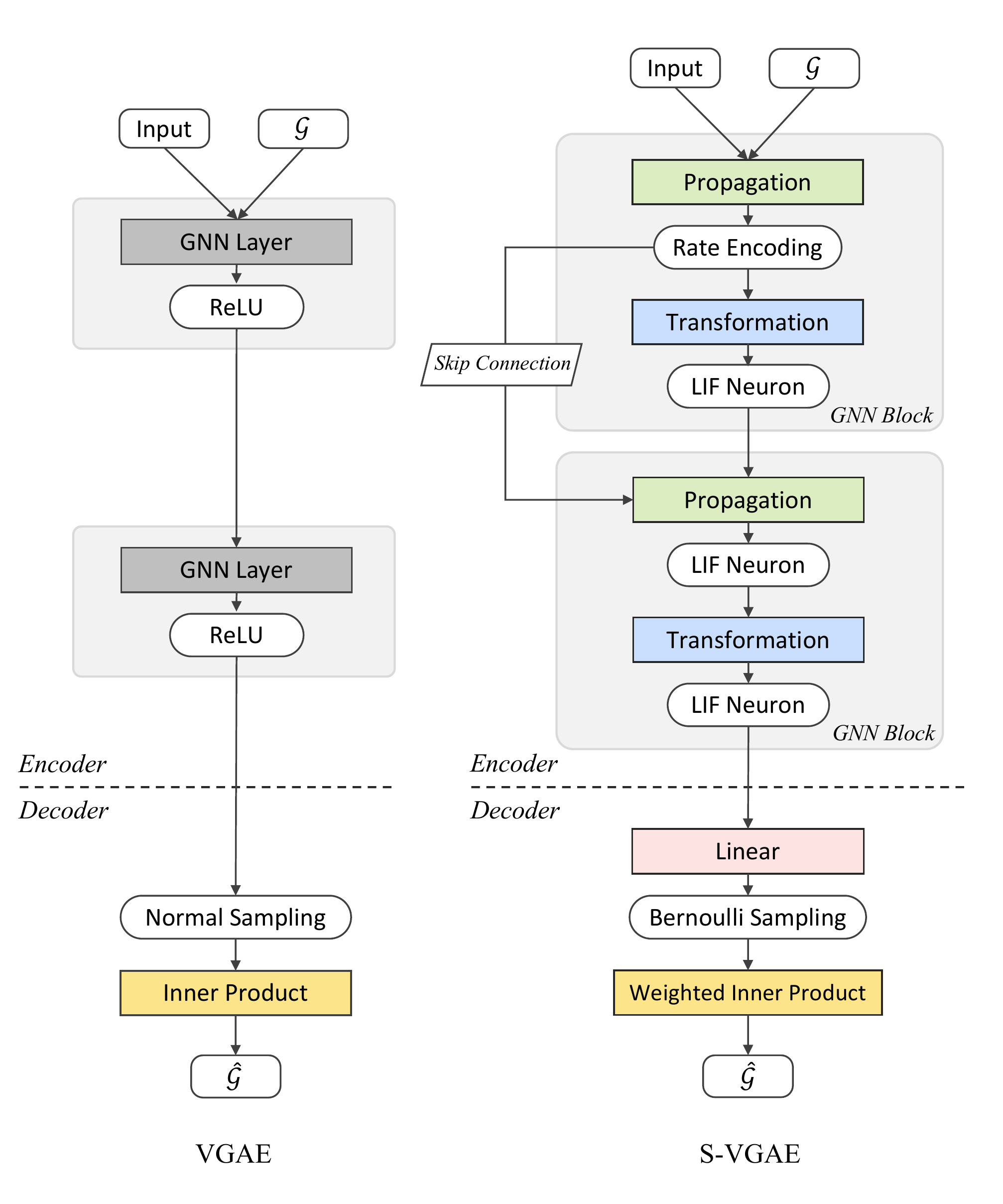}}\par
		\caption{Architecture design of VGAE (left) and our proposed S-VGAE (right).}
		\label{svgae}
	\end{center}
\end{figure}

Based on the VGAE \cite{kipf2016variational} architecture, the proposed S-VGAE includes a spiking GNN encoder to learn the latent structure of graphs and a probabilistic spiking decoder for graph reconstruction, as shown in Fig.~\ref{svgae}. Specifically, the node features are first transformed to $T$-step spiking signals $\X^{1:T}\in\R^{N\times C_0\times T}$ using the Poisson rate encoding scheme \cite{diehl2015fast}, and are fed into the encoder sequentially by time steps. Then, with the spiking outputs of the encoder $\HH^{1:T}\in\R^{N\times C_L\times T}$, where $C_L$ is the dimension of feature channels, the decoder samples stochastic binary signals from variational posteriors, which are subsequently input to the readout layer for graph reconstruction.

\paragraph{Spiking GNN Encoder}
Following conventional VGAEs, the encoder could be parameterized by GNN layers, i.e.
\begin{align}\label{GNN}
	\HH^{t,l+1}=\HH^{t,l}\ast_{\G}\g^{l}_{\theta}\W^{l},
\end{align}
where $\HH^{t,l}\in\R^{N\times C_l}$ is the output spiking signals at $t$ step of the $l$-th GNN block and $\HH^{t,0}=\X^t$, $\W^l\in\R^{C_{l+1}\times C_l}$ is a trainable synaptic weight matrix, and $\g^l_{\theta}$ denotes the propagation aggregator with coefficients $\theta$. However, Eq.~(\ref{GNN}) is not bio-plausible and can greatly increase the energy consumption of the encoder. To be more specific, a GNN aggregator can be divided into two independent layers, namely the propagation and transformation. Since both the propagation aggregator coefficients $\theta$ (e.g., the normalized Laplacian in GCNs) and the transformation weights $\W^l$ are floating-point numbers, unifying these two operations in one layer as in conventional GNNs will cause floating-point MACs, which are inefficient to implement on the neuromorphic hardware. Based on these observations, we propose the spiking GNN block as a stack of two layers to separately perform propagation and transformation, both of which employ spiking neurons to output discrete signals with the spatio-temporal dynamics. Notably, the decoupling of these two operations in GNNs has been adopted by previous ANN-based methods \cite{wu2019simplifying, liu2020towards}, but with very different motivations. In this paper, we consider this strategy for the sake of adjusting GNNs to the SNN framework, which is a novel idea for the both communities.

Despite the superiority of bio-fidelity and energy efficiency, the additional spiking neurons between the propagation and transformation layers may exaggerate the difficulty in building a deep SNN architecture to learn the high-order neighbor information. For some large graphs, it is necessary to aggregate the features of higher-order neighbors from more than one-hop away, especially when the graph is very sparse and the one-hop neighbor information is insufficient. Therefore, we add the skip-connections between propagation layers for a hierarchical encoder architecture with multiple GNN blocks. Formally, the proposed spiking GNN encoder is given as
\begin{align}
	&\tilde{\HH}^{t,l+1}=\Phi([\HH^{t,l}\vert\tilde{\HH}^{t,l}]\ast_{\G}\g^{l}_{\theta}),\label{propagation}\\
	&\HH^{t,l+1}=\Phi(\tilde{\HH}^{t,l+1}\W^{l}),\label{transformation}
\end{align}
where $[\cdot\vert\cdot]$ denotes the matrix concatenation and $\Phi(\cdot)$ is a spiking neuron. The decoupling of propagation and transformation operations guarantees that each layer of the encoder emits spiking signals and transforms the floating-point MACs into ACs. Moreover, the skip-connections enable features from the propagation layer $\tilde{\HH}^{t,l}$ to be directly passed to the next block by skipping the additional spiking neurons in the transformation layer. Such skip-connections can reduce the unavoidable information loss for the spiking neurons to map continuous natural signals into discrete spiking signals, referred to as the quantization error \cite{bu2022optimal}.

\paragraph{Model Instantiating: Spiking GCN Encoder}
Following VGAEs, the proposed S-VGAE employs the GCN \cite{kipf2017semi} aggregator as an example. Combining with Eq.~(\ref{propagation}), the spiking GCN propagation is formulated as
\begin{align}\label{GCN}
	\HH^{t,l}=\Phi(\tilde{\D}^{-\frac{1}{2}}\tilde{\A}\tilde{\D}^{-\frac{1}{2}}[\HH^{t,l}\vert\tilde{\HH}^{t,l}]),
\end{align}
where $\tilde{\A}=\A+\I_N$ is the adjacency matrix with added loops, i.e., the approximated Laplacian, and $\I_N$ is the $N$-dimensional identity matrix. $\tilde{\D}$ is the diagonal matrix of degrees with non-zero elements being the row (or column) summations of $\tilde{\A}$. Note that the proposed encoder framework can also be applied to other general GNNs, such as GAT \cite{velivckovic2018graph} and GraphSAGE \cite{hamilton2017inductive}.

\paragraph{Probabilistic Spiking Decoder}
The decoder consists of a linear layer with probabilistic spiking neurons to sample latent variables and a W-IP readout layer to reconstruct the graph using the binary latent variables. Following conventional VAE-based methods, we consider the mean-field approximation \cite{kingma2014auto} to factorize the posterior as
\begin{align}\label{posterior}
	q(\Z^{1:T}\vert\X^{1:T})=\prod_{n=1}^{N}\prod_{t=1}^{T}q(\z_n^{t}\vert\x_n^{1:t},\z_n^{1:(t-1)}),
\end{align}
where $\Z^{1:T}=(\z_1^{1:T},\dots,\z_N^{1:T})'\in\R^{N\times C_z\times T}$ are the latent variables. In particular, as $\z_n^{t}$ must be dynamic discrete signals, the common Normal prior is no longer appropriate for SNNs. Consequently, we employ the Bernoulli process to sample spiking latent variables, i.e., $\z_n^{t}\sim\Ber(\pi)$, where $\pi\in(0,1)$ is the prior probability for firing. Typically, we expect a low firing rate since a large number of spikes may lead to long latency and high energy cost \cite{stockl2021optimized, davidson2021comparison}. Thus, we set a small $\pi$ to generate sparse binary latent variables.

Since $\z_n^{t}$ are binary spiking signals, it is unfeasible to compute the IP of latent variables to reconstruct graphs as in VGAE. Hence, we propose the W-IP readout layer, defined as
\begin{align}\label{readout}
	\eta_{nm}^t=\w\odot{\z_n^{t}}^{\top}\z_m^{t},
\end{align}
where $\eta_{nm}^t$ is the log-odds (logit) to form an edge between node $n$ and $m$, $\odot$ denotes the element-wise vector multiplication, and $\w\in\R^{C_z}$ is a trainable weight vector. Note that Eq.~(\ref{readout}) is also energy-efficient with only AC operations.

Finally, $\eta_{nm}^{1:T}$ are decoded to edges $y_{nm}$ using non-firing spiking neurons, derived as
\begin{align}\label{decode}
	\PP(y_{nm}=1\vert\z_n^{t},\z_m^{t})=\sigma\left(\sum_{t=1}^{T}\tau_{out}^{T-t}\eta_{nm}^{t}\right),
\end{align}
where $\sigma(\cdot)$ denotes the sigmoid function and $\tau_{out}$ is the decay constant of the readout neuron.

\subsection{Spiking Neuron Models}

Following the recent progress of SNNs \cite{kamata2022fully, zhu2022spiking}, we adopt the iterative LIF model \cite{wu2019direct} with soft reset for the spiking neurons, formulated as
\begin{align}\label{LIF}
	&u_{n,c}^t=\tau(u_{n,c}^{t-1}-o_{n,c}^{t-1}V_{th})+x_{n,c}^t,\\
	&o_{n,c}^{t}=\text{F}(u_{n,c}^t-V_{th}),
\end{align}
where $u_{n,c}^t$ is the membrane potential for the $c$-th feature of node $n$ at time step $t$, $\tau$ is the decay constant, $V_{th}$ is the firing threshold, $x_{n,c}^t$ and $o_{n,c}^t$ are the pre-synaptic input and post-synaptic output of the neuron, given in Eq.~(\ref{propagation}) and (\ref{transformation}). $\text{F}(\cdot)$ is the firing function, which determines the scheme to transform continuous membrane potential into discrete spiking signals. In order to leverage the robustness of VAEs via random sampling, as well as the efficiency of surrogate gradient-based training, we adopt deterministic neurons for the spiking GNN layers of the encoder and propose the probabilistic LIF neurons with a soft threshold for the linear layer of the decoder, respectively.

\paragraph{Deterministic LIF Model} For the encoder layers, we leverage the Heaviside step function as $\text{F}(\cdot)$, the value of which is 1 when $u_{n,c}\geq V_{th}$ and 0 otherwise. For training with backpropagation, we leverage the rectangular function as the surrogate gradient \cite{wu2018spatio}, i.e.,
\begin{align}\label{rectangular}
	\frac{\partial o_{n,c}^{t}}{\partial u_{n,c}^t}\approx\frac{1}{a}\sign(\vert u_{n,c}^t-V_{th}\vert<\frac{a}{2}),
\end{align}
where $\sign(\cdot)$ is the signum function and $a$ determines the peak width. In particular, for the first layer ($l=0$) of propagation, we leverage the Poisson rate encoding instead of LIF model since it requires no weight and is essentially equivalent to a pre-processing step of spiking features.

\paragraph{Probabilistic LIF Model} Inspired by the spike response model (SRM) with stochastic threshold \cite{jang2019introduction}, we propose the probabilistic LIF model to generate binary latent variables $z_{n,c}^t$ from a Bernoulli Process, i.e.,
\begin{align}\label{bernoulli}
	\PP(z_{n,c}^t=1\vert u_{n,c}^t)=\sigma(u_{n,c}^t-V_{th}).
\end{align}
Conditioned on the spiking history, the difference between the membrane potential $u_{n,c}^t$ and threshold $V_{th}$ serves as the log-odds for a neuron to emit signals, which can improve model robustness via the soft firing threshold. For training, we use the sigmoid surrogate gradient, i.e., $\partial z_{n,c}^{t}/\partial u_{n,c}^t\approx\sigma'(u_{n,c}-V_{th})$. Note that Eq.~(\ref{bernoulli}) can also be performed using the autoregressive Bernoulli spike sampling method \cite{kamata2022fully}, which is more feasible to implement on the neuromorphic hardware.

\subsection{Loss Function}

Following the classic VAEs \cite{kingma2014auto}, we use the negative evidence lower bound (ELBO) as the loss function, given as
\begin{align}\label{loss}
	\mathcal{L}=&-\sum_{n=1}^N\sum_{m=1}^N\mathbb{E}_q[\log\thinspace p(y_{nm}\vert\z_n^{1:T},\z_m^{1:T})]\nonumber\\
	            &+\frac{1}{T}\sum_{n=1}^N\sum_{t=1}^T\KL[q(\z_n^{t}\vert\X^{1:t})\Vert p(\z_n^{t})].
\end{align}
The first term is the negative cross-entropy loss of graph reconstruction. The second term is the Kullback-Leibler (KL) divergence between the variational posterior $q(\cdot)$ and the Bernoulli prior $p(\cdot)$, which serves as a regularization term to adjust the sparsity of the binary latent variables $\z_{n}^t$. The full expression of the KL divergence and a pseudo-code for training our method is given in Appendix.

\section{Experiments}

We conduct extensive experiments on multiple benchmark graph datasets to evaluate the energy efficiency as well as the link prediction performances of our proposed method, compared with other competitive ANN- and SNN-based graph representation learning methods.

\subsection{Datasets}

The experiments are conducted on five real-world graph datasets, including three medium-sized citation network datasets, i.e., Cora, CiteSeer and PubMed \cite{sen2008collective}, and a large-scale open graph benchmark (OGB) dataset, i.e., ogbl-collab \cite{wang2020microsoft}. In the citation networks, nodes and edges represent papers and citations, respectively. The Cora and PubMed datasets contain bag-of-words features, and the CiteSeer dataset includes one-hot category labels. The ogbl-collab dataset is a collaboration network where nodes represent authors and edges denote collaborations between them. Each node comes with continuous word embeddings of the papers published by the author, which are transformed to spiking node features via the Poisson rate encoding for SNN-based methods. All edges are randomly split as 85\% for training, 10\% for testing, and 5\% for validation. The descriptive statistics of these datasets are presented in Table~\ref{datasets}.

\begin{table}
	\centering
	\resizebox{\columnwidth}{!}{
		\begin{tabular}{lcccc}
			\toprule
			Dataset    &\#Nodes &\#Edges    &\#Features & Density\\
			\midrule
			Cora       & 2,708 &     5,278& 1,433      &0.072\%\\
			CiteSeer   & 3,312 &     4,552& 3,703      &0.042\%\\
			PubMed     &19,717 &    44,324& 500        &0.011\%\\
			ogbl-collab&235,868& 1,285,465& 128        &0.002\%\\
			\bottomrule
		\end{tabular}
	}
	\caption{Descriptive statistics of graph datasets.}\label{datasets}
\end{table}

\begin{table*}
	\centering
	\resizebox{\textwidth}{!}{
		\begin{tabular}{lc|cccc|cc|cc}
			\toprule
			\rule{0pt}{10pt}&&\multicolumn{4}{c|}{\underline{\textbf{ANN-Based}}}&\multicolumn{2}{c|}{\underline{\textbf{SNN-Based}}}&\multicolumn{2}{c}{\underline{\textbf{Our Methods}}}\\[1pt]
			\multicolumn{2}{c|}{\textbf{Performance}}&GCN*&GAT*&VGAE&Graphite&GC-SNN*&SpikingGCN*&S-GAE&S-VGAE\\
			\midrule
			\rule{0pt}{9pt}\multirow{2}{*}{\textbf{Cora}}&AUC&91.2 $\pm$ 0.7&89.6 $\pm$ 0.5&\underline{91.4 $\pm$ 0.0}&89.5 $\pm$ 0.9&90.8 $\pm$ 0.7&\underline{91.4 $\pm$ 0.1}&\underline{91.4 $\pm$ 0.8}&\textbf{92.6 $\pm$ 0.5}\\[1pt]
			&AP&91.2 $\pm$ 0.6&89.6 $\pm$ 0.7&\textbf{92.6 $\pm$ 0.0}&90.7 $\pm$ 0.5&90.1 $\pm$ 1.0&90.7 $\pm$ 0.6&90.9 $\pm$ 1.0&\underline{92.2 $\pm$ 0.6}\\
			\midrule
			\rule{0pt}{9pt}\multirow{2}{*}{\textbf{CiteSeer}}&AUC&90.5 $\pm$ 0.5&91.0 $\pm$ 1.0&90.8 $\pm$ 0.0&\underline{91.5 $\pm$ 0.9}&89.6 $\pm$ 0.8&87.8 $\pm$ 1.3&91.8 $\pm$ 0.7&\textbf{92.1 $\pm$ 0.9}\\[1pt]
			&AP&91.9 $\pm$ 0.5&92.4 $\pm$ 1.0&92.0 $\pm$ 0.0&91.5 $\pm$ 0.9&91.4 $\pm$ 0.5&91.1 $\pm$ 1.0&\textbf{92.7 $\pm$ 0.5}&\underline{92.5 $\pm$ 0.7}\\
			\midrule
			\rule{0pt}{9pt}\multirow{2}{*}{\textbf{PubMed}}&AUC&95.7 $\pm$ 0.2&93.1 $\pm$ 0.2&94.4 $\pm$ 0.0&\underline{96.1 $\pm$ 0.1}&\textbf{96.6 $\pm$ 0.1}&92.1 $\pm$ 0.2&96.2 $\pm$ 0.1&95.9 $\pm$ 0.2\\[1pt]
			&AP&96.1 $\pm$ 0.2&93.0 $\pm$ 0.3&94.7 $\pm$ 0.0&\textbf{96.3 $\pm$ 0.1}&\textbf{96.3 $\pm$ 0.1}&92.3 $\pm$ 0.3&95.9$\pm$ 0.1&95.6 $\pm$ 0.2\\
			\midrule
			\rule{0pt}{9pt}\multirow{2}{*}{\textbf{ogbl-collab}}&AUC&94.1 $\pm$ 0.1&92.2 $\pm$ 0.5&\textbf{95.0 $\pm$ 0.1}&94.8 $\pm$ 0.2&94.6 $\pm$ 0.3&90.8 $\pm$ 0.1&94.8 $\pm$ 0.3&\underline{94.9 $\pm$ 0.4}\\[1pt]
			&AP&94.8 $\pm$ 0.2&92.6  $\pm$ 0.5&\textbf{95.9 $\pm$ 0.1}&95.4 $\pm$ 0.3&95.0 $\pm$ 0.2&91.5 $\pm$ 0.2&95.5 $\pm$ 0.4&\underline{95.6 $\pm$ 0.5}\\
			\midrule\midrule&&&&&&&&&\\[-8pt]
			\multicolumn{2}{c|}{\textbf{Energy Cost}}&GCN*&GAT*&VGAE&Graphite&GC-SNN*&SpikingGCN*&S-GAE&S-VGAE\\
			\midrule
			\rule{0pt}{9pt}\multirow{2}{*}{\textbf{Cora}}&$\text{E}^{\text{F}}$&88.68&265.17&92.70&172.99&29.76&38.95&\textbf{4.23}&\textbf{4.23}\\[1pt]
			&$\text{E}^{\text{I}}$&61.69&184.46&64.49&119.91&20.11&25.95&\underline{0.50}&\textbf{0.47}\\
			\midrule
			\rule{0pt}{9pt}\multirow{2}{*}{\textbf{CiteSeer}}&$\text{E}^{\text{F}}$&222.22&444.10&226.18&319.98&36.85&44.78&\underline{5.82}&\textbf{5.80}\\[1pt]
			&$\text{E}^{\text{I}}$&154.59&308.94&157.34&222.38&\underline{25.18}&29.10&\underline{0.71}&\textbf{0.65}\\
			\midrule
			\rule{0pt}{9pt}\multirow{2}{*}{\textbf{PubMed}}&$\text{E}^{\text{F}}$&33.80&100.38&37.86&337.18&49.96&21.38&\textbf{7.70}&\textbf{7.70}\\[1pt]
			&$\text{E}^{\text{I}}$&23.52&69.82&26.33&234.13&34.09&14.76&\textbf{0.86}&\textbf{0.86}\\
			\midrule
			\rule{0pt}{9pt}\multirow{2}{*}{\textbf{ogbl-collab}}&$\text{E}^{\text{F}}$&32.79&822.49&49.16&869.28&39.64&16.85&\textbf{10.58}&\textbf{10.58}\\[1pt]
			&$\text{E}^{\text{I}}$&22.81&572.17&34.20&602.99&24.60&10.51&\textbf{1.16}&\underline{1.18}\\
			\bottomrule
		\end{tabular}
	}
	\caption{Experimental results on link prediction in terms of AUC and AP scores (in \%) and the energy cost of floating-point ($\text{E}^{\text{F}}$) and integer ($\text{E}^{\text{I}}$) MAC operations ($\times$10$^4$pJ). The best results are in bold and the second are underlined. The asterisk * indicates that the method has been modified using the GAE framework.}\label{results}
\end{table*}

\subsection{Baselines}

We compare the proposed methods with both ANN-based and SNN-based methods for graph representation learning. The ANN-based methods include two classic GNN frameworks, i.e. GCN \cite{kipf2017semi} and GAT \cite{velivckovic2018graph}, and two VAE-based methods, i.e. VGAE \cite{kipf2016variational} and Graphite \cite{grover2019graphite}, of which VGAE can be regarded as the ANN-based counterpart of our proposed S-VGAE. We also consider two recent SNN-based methods, i.e. GC-SNN \cite{xu2021exploiting} and SpikingGCN \cite{zhu2022spiking}. Note that all GNN-based methods, including GCN, GAT, GC-SNN and SpikingGCN, are incapable of addressing the link prediction task, because these above methods can only produce the representation of a single node. To solve this problem, we leverage the graph auto-encoder (GAE) framework by adding a decoder with the IP layer for the ANN-based methods and the proposed W-IP layer for the SNN-based methods. In addition, we also consider a deterministic variant of our proposed S-VGAE, which we refer to as S-GAE, by substituting the ordinary LIF model for the proposed probabilistic LIF model in the decoder.

\subsection{Experimental Settings}

For the proposed S-GAE and S-VGAE methods, we set the encoder to have one (for Cora and CiteSeer) or two (for PubMed and obgl-collab) spiking GCN blocks with 64 neurons in each layer, and the probabilistic decoder to have one spiking linear layer with 64 stochastic LIF neurons. The hyperparameter settings are as follows: the time window $T=10$, the firing threshold $V_{th}=0.2$, the decay constant $\tau=0.25$, $\tau_{out}=0.8$, and the Bernoulli prior probability $\pi=0.1$. For all datasets, our model is trained with less than 500 epochs and a learning rate from 0.001 to 0.05.

The comparative baselines are set to have similar architectures with the same layer sizes for fair comparisons. Specifically, the GNN-based methods are set to have two aggregation layers and one IP readout layer. The VAE-based methods are comprised of an encoder with two GCN layers, and a probabilistic decoder with an IP layer for graph reconstruction using Normal latent variables. The SNN-based methods are implemented under the original architecture except for the readout layer. Other hyperparameters are set as the default. All methods are implemented on 2080 Ti GPUs with 64 GiB RAM.

\subsection{Evaluation Metrics}
For benchmark comparisons, we employ the area under the ROC curve (AUC) and average precision (AP) as the performance metrics of link prediction. To investigate the energy efficiency of different methods, following \citet{kim2020spiking}, we calculate the energy cost of floating-point $\text{E}^{\text{F}}$ and integer $\text{E}^{\text{I}}$ operations as the summations of AC and multiply (MUL) energy consumption, i.e.,
\begin{align}
	\text{E}^{\text{F}}&=\text{e}_{\AC}^{\text{F}}\times\text{N}_{\AC}+\text{e}_{\MUL}^{\text{F}}\times\text{N}_{\MUL},\\
	\text{E}^{\text{I}}&=\text{e}_{\AC}^{\text{I}}\times\text{N}_{\AC}+\text{e}_{\MUL}^{\text{I}}\times\text{N}_{\MUL},
\end{align}
where $\text{e}_{\AC}^{\text{F}}$ ($\text{e}_{\AC}^{\text{I}}$) and $\text{e}_{\MUL}^{\text{F}}$ ($\text{e}_{\MUL}^{\text{I}}$) denote the unit energy consumption of AC and MUL operations for floating points (integers), $\text{N}_{\AC}$ and $\text{N}_{\MUL}$ denote the number of AC and MUL operations, respectively. According to \citet{horowitz20141}, $\text{e}_{\AC}^{\text{F}}$=0.9pJ, $\text{e}_{\MUL}^{\text{F}}$=3.7pJ, $\text{e}_{\AC}^{\text{I}}$=0.1pJ, and $\text{e}_{\MUL}^{\text{I}}$=3.7pJ, where 1pJ$\approx$2.78$\times$10$^{-7}$kW$\cdot$h. The detailed calculation formulas of $\text{N}_{\AC}$ and $\text{N}_{\MUL}$ for different layers in our methods are given in Appendix. For a fair comparison across different datasets, we only report the average energy consumption of predicting a single link.

\subsection{Quantitative Results}

\paragraph{Link Prediction Performances}
The experimental results of link prediction are presented in Table~\ref{results}, where all reported results are obtained using the means and standard deviations of 10 independent runs with different random seeds. For all datasets, the proposed S-VGAE performs better than or comparably with the ANN-based methods, which verifies the effectiveness of the proposed W-IP layer for link prediction using spiking node representations. Furthermore, S-VGAE also achieves superior results compared with other SNN-based methods (except for the GC-SNN method on the PubMed dataset), because the proposed spiking encoder with multiple GNN blocks and skip-connections can learn the higher-order neighbor information.


\paragraph{Energy Efficiency Analysis}

\begin{figure*}[!htbp]
	\centering
	\subfloat[Cora]{
		\includegraphics[width=0.24\textwidth]{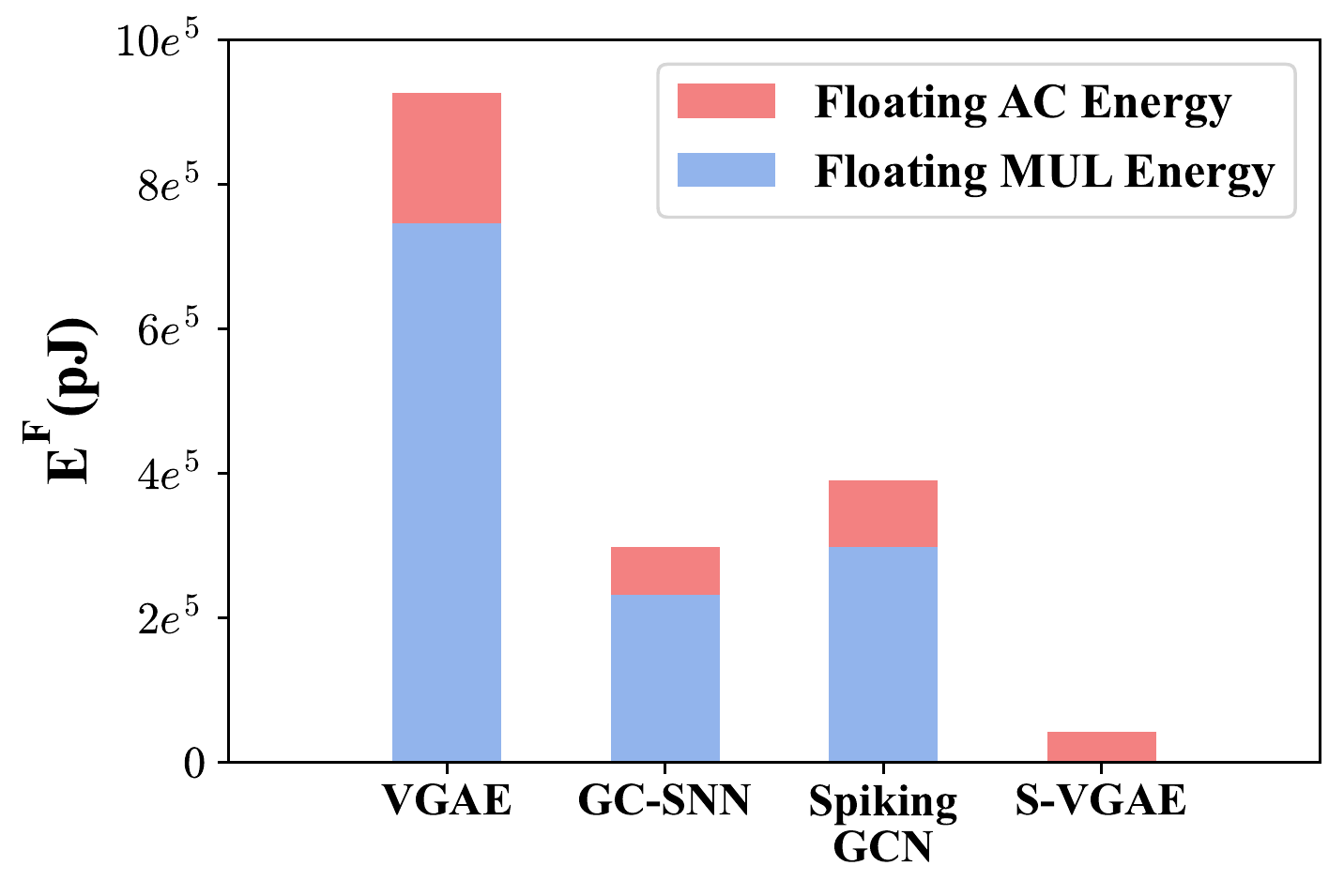}
	}
	\subfloat[CiteSeer]{
		\includegraphics[width=0.24\textwidth]{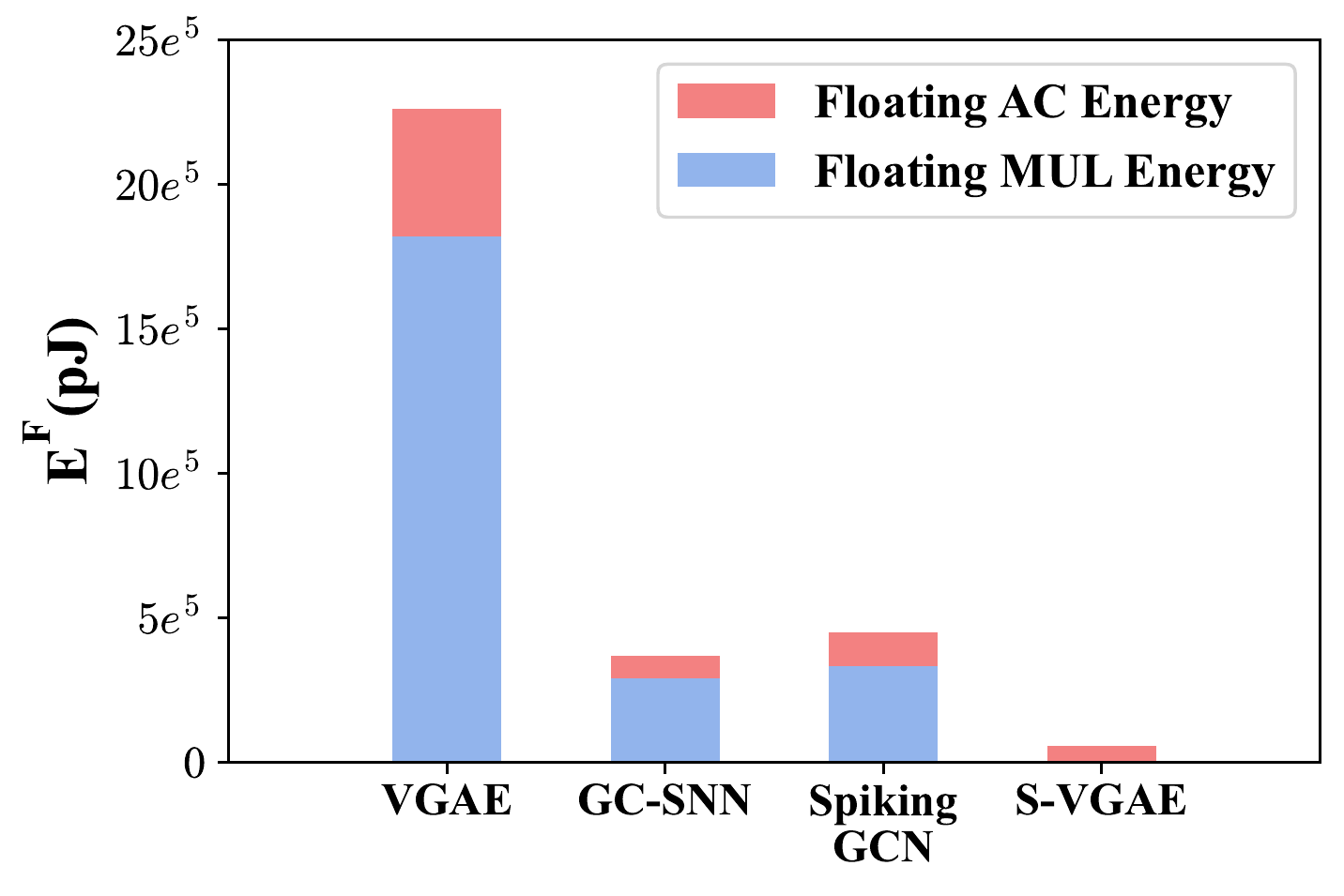}
	}
	\subfloat[PubMed]{
		\includegraphics[width=0.24\textwidth]{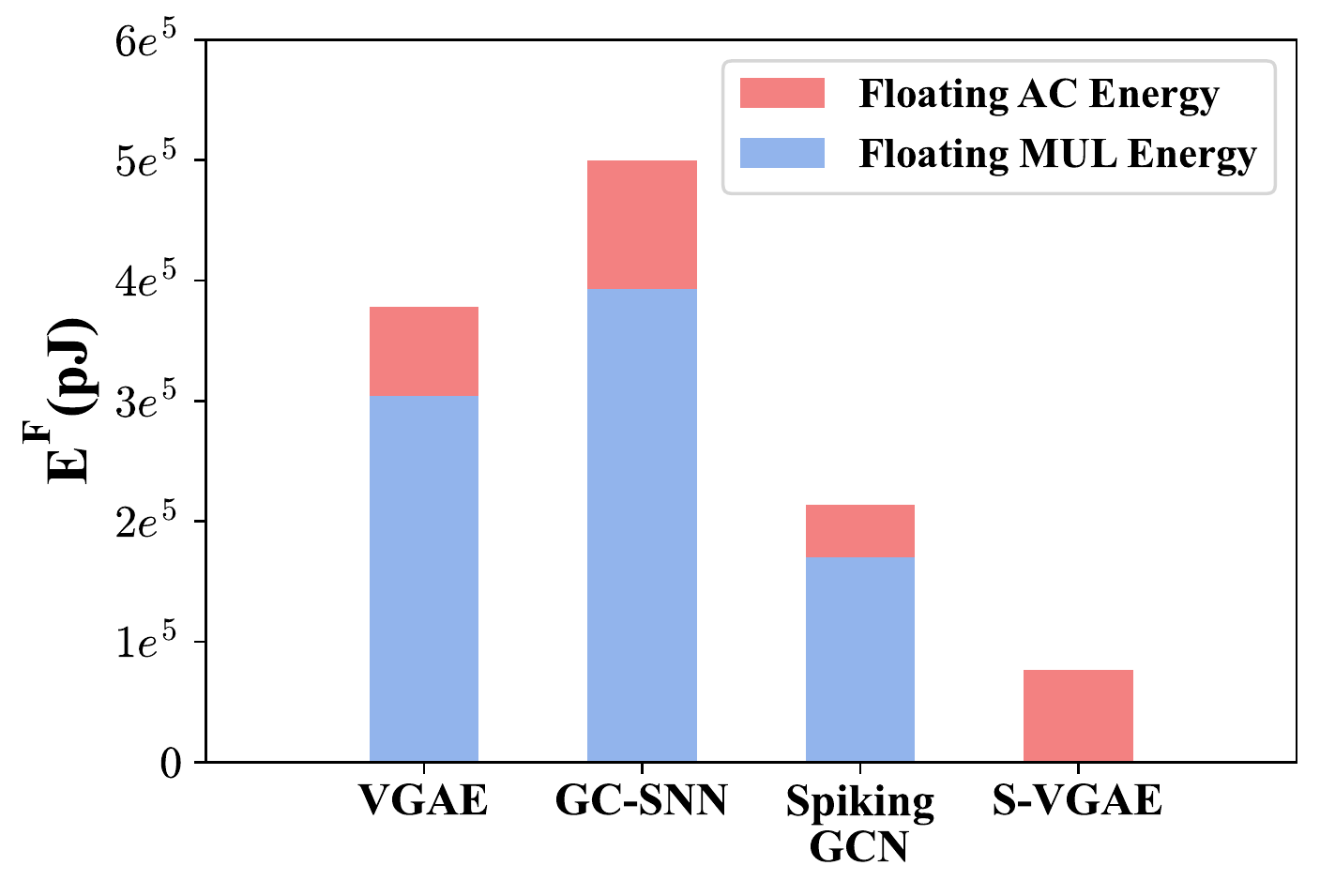}
	}
	\subfloat[ogbl-collab]{
		\includegraphics[width=0.24\textwidth]{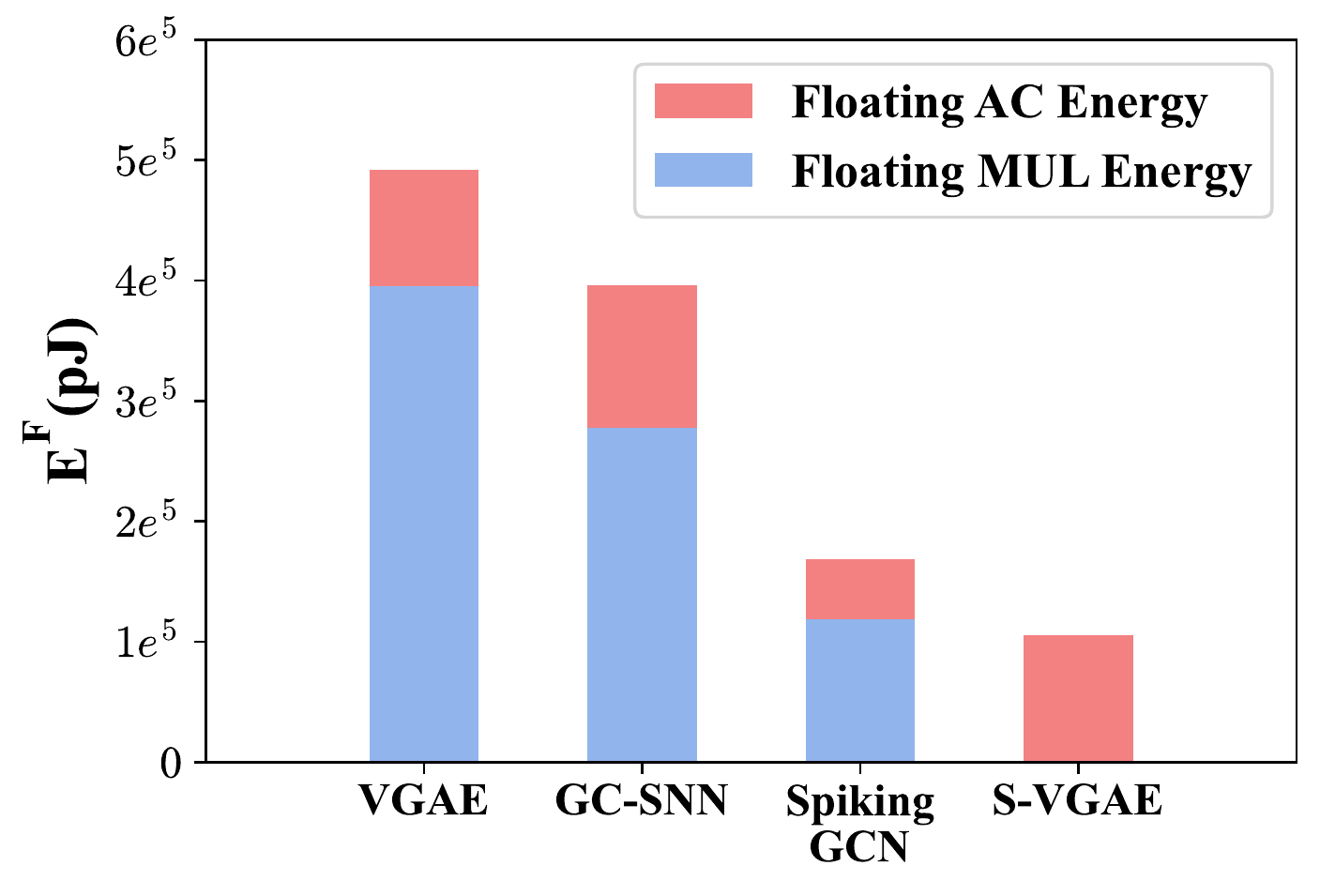}
	}
	
	\subfloat[Cora]{
		\includegraphics[width=0.24\textwidth]{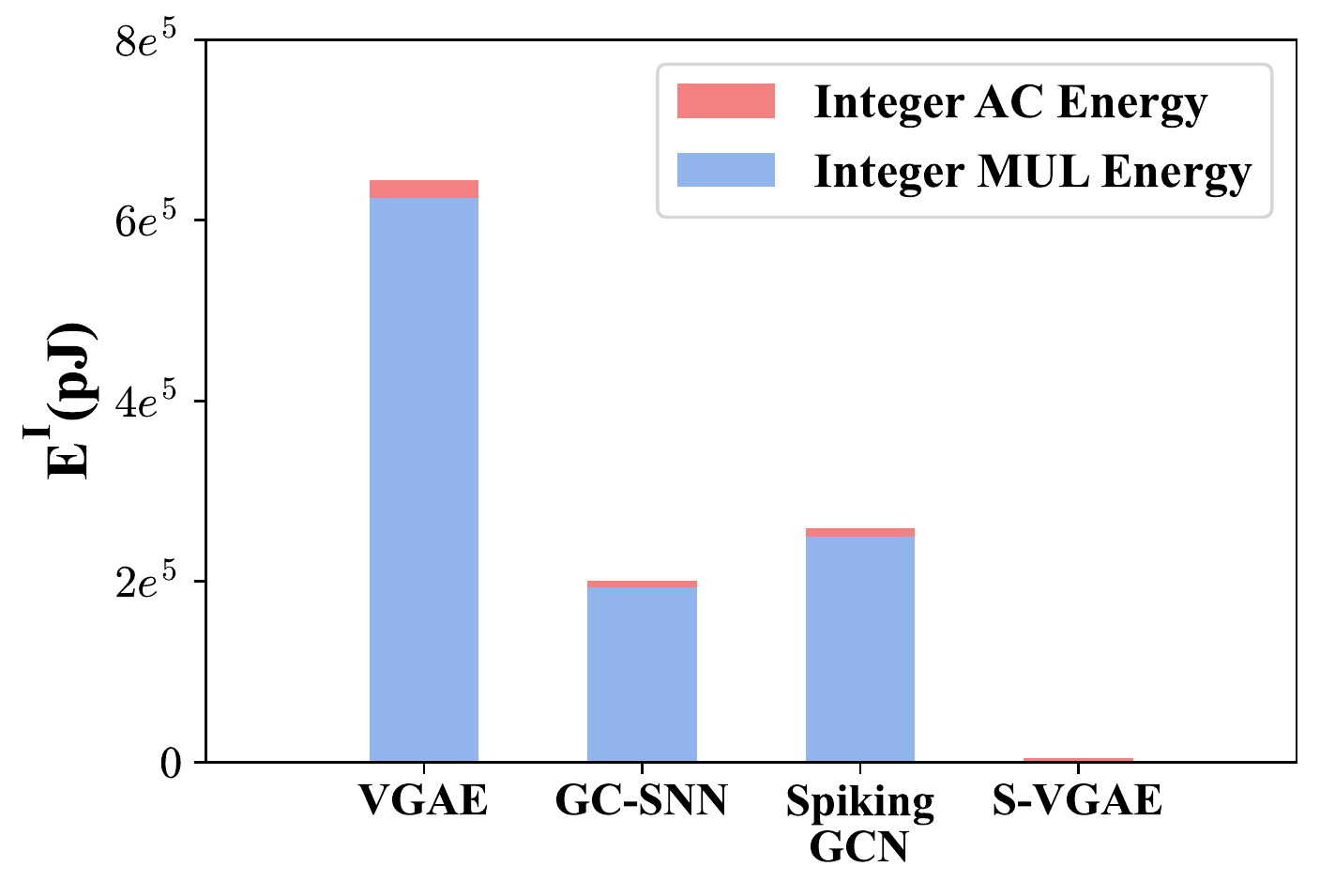}
	}
	\subfloat[CiteSeer]{
		\includegraphics[width=0.24\textwidth]{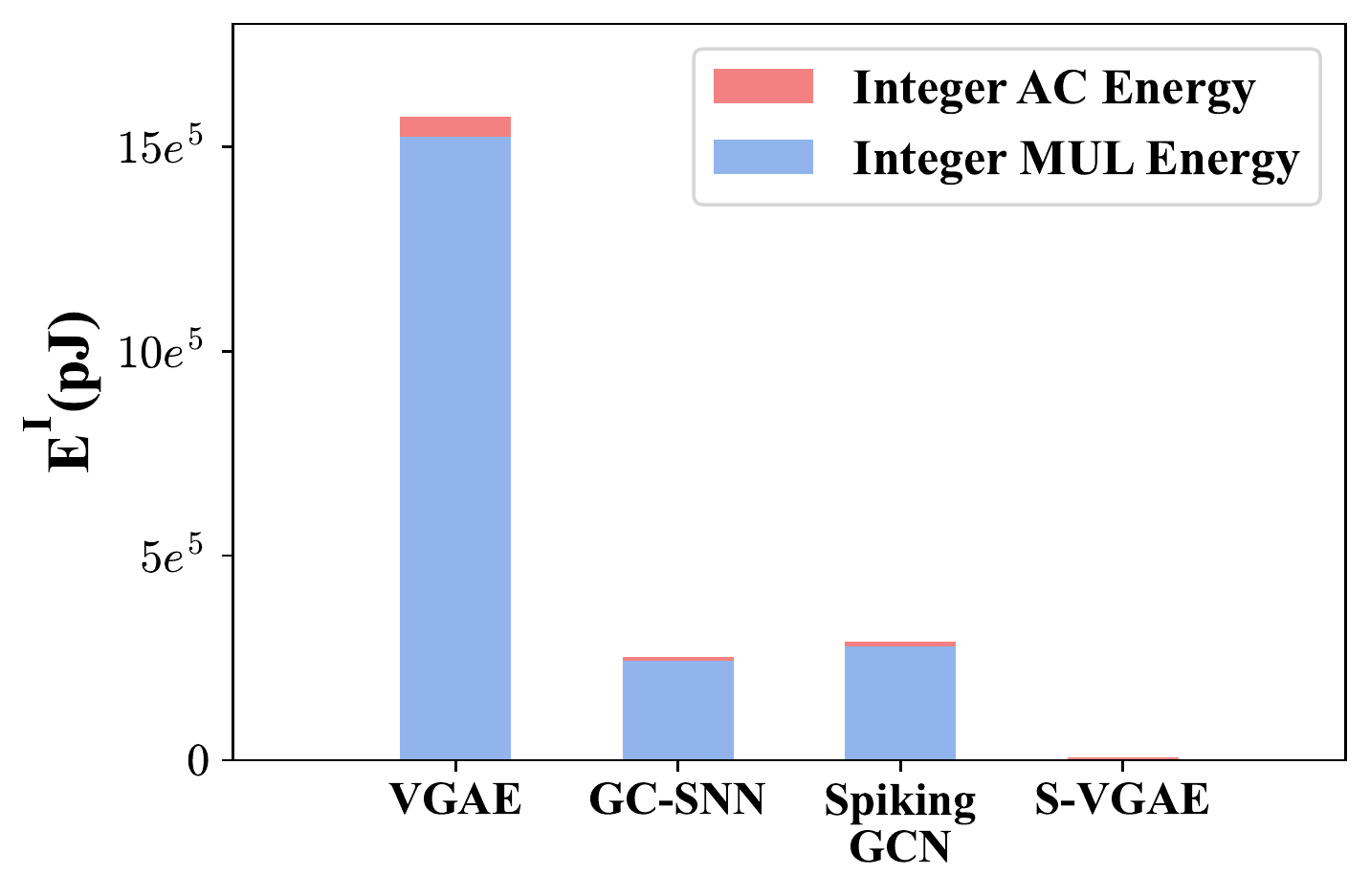}
	}
	\subfloat[PubMed]{
		\includegraphics[width=0.24\textwidth]{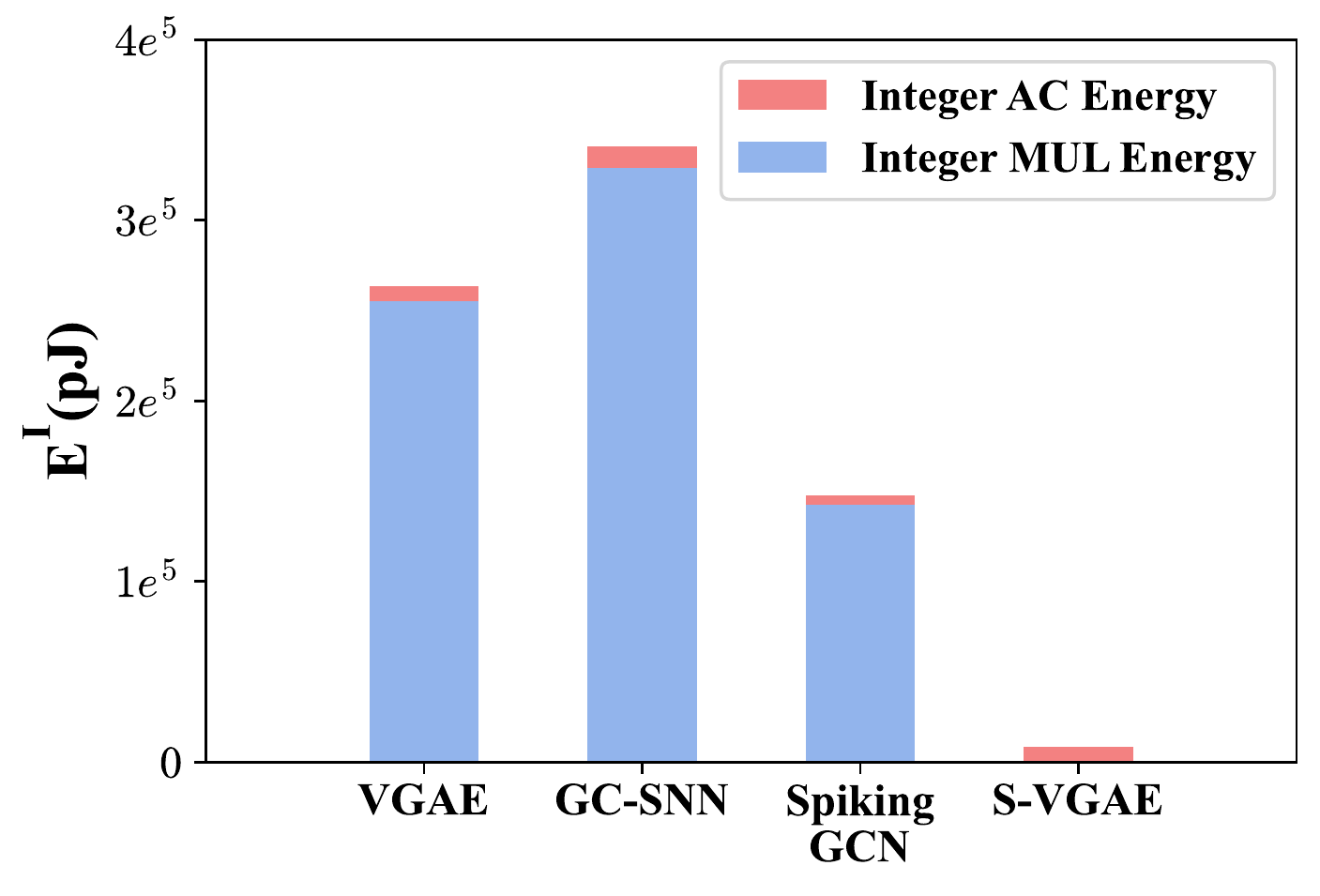}
	}
	\subfloat[ogbl-collab]{
		\includegraphics[width=0.24\textwidth]{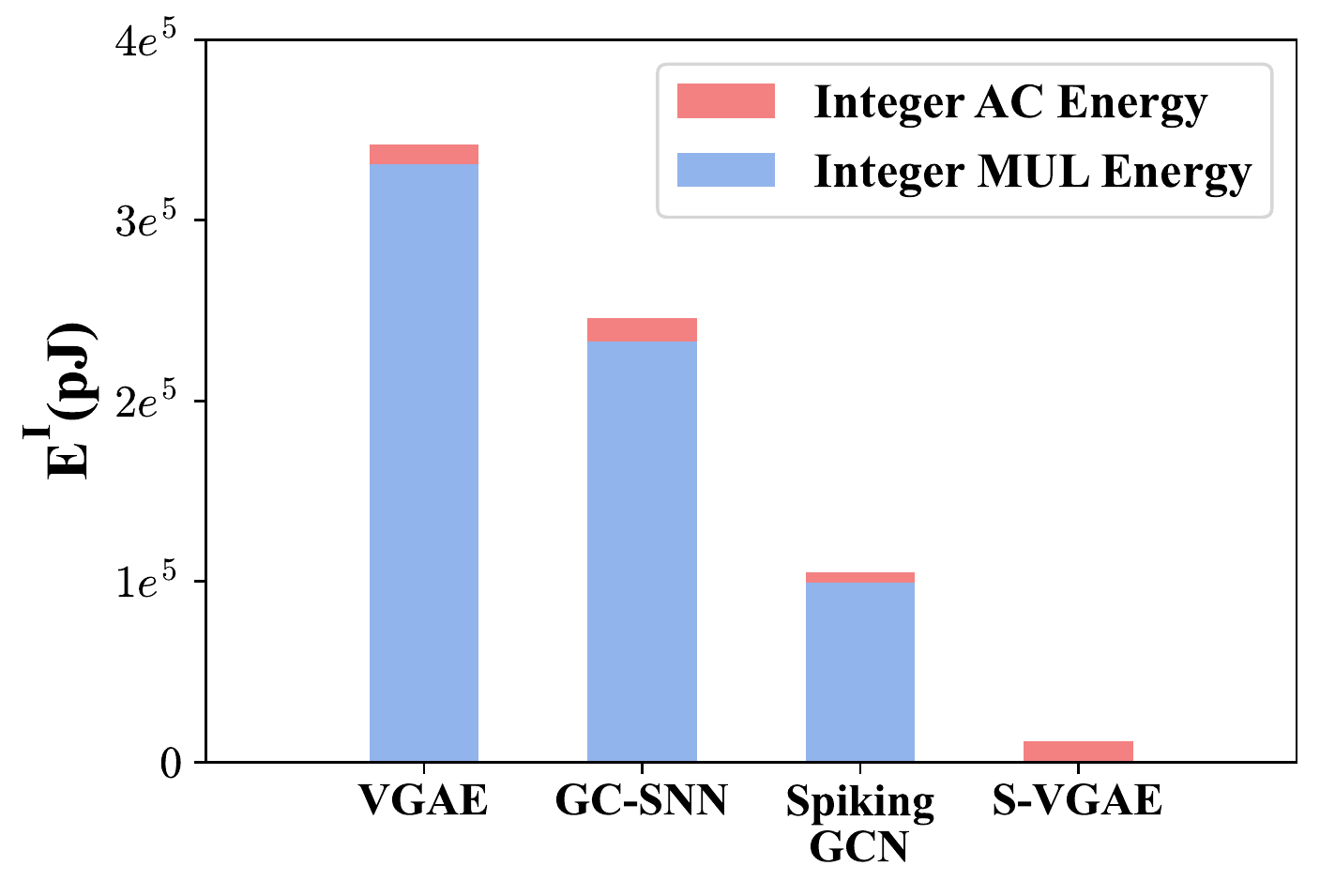}
	}
	\centering
	\caption{Comparative results on energy consumption of floating-point operations $\text{E}^{\text{F}}$ (a-d) and integer operations $\text{E}^{\text{I}}$ (e-h) to predict a single link on average. The red and blue colors stand for accumulate (AC) and multiply (MUL) operations, respectively.}\label{energy}
\end{figure*}

Table~\ref{results} also presents the average energy consumption of predicting per link for different methods. The experimental results demonstrate that our proposed methods significantly outperform all other methods on energy efficiency. For a more intuitive presentation, in Fig.~\ref{energy}, we also illustrate the energy consumption of our proposed S-VGAE and those of other SNN-based methods GC-SNN and SpikingGCN, as well as an ANN-based counterpart VGAE. The results show that, for all compararing methods, the vast majority of energy consumption comes from the MUL operations, of which the unit energy cost is much higher than ACs (about 4.1 times over ACs for floating points and 31 times for integers). On the contrary, the proposed S-VGAE method nearly has zero MUL operations and thus consumes considerably lower energy through decoupling the GNN blocks into spiking propagation and transformation layers.


\begin{table}[htbp]
	\centering
	\resizebox{\columnwidth}{!}{
		\begin{tabular}{lcccc}
			\toprule
			\rule{0pt}{10pt}&\textbf{Cora}&\textbf{CiteSeer}&\textbf{PubMed}&\textbf{ogbl-collab}\\
			\midrule
			\rule{0pt}{10pt}One-Block&\textbf{92.6 $\pm$ 0.5}&\textbf{92.1 $\pm$ 0.9}&93.3 $\pm$ 0.2&89.8 $\pm$ 1.0\\[2pt]
			Two-Block&90.8 $\pm$ 2.3&84.5 $\pm$ 0.5&\textbf{95.9 $\pm$ 0.2}&\textbf{94.9 $\pm$ 0.4}\\[2pt]
			\quad w/o. Skip.&83.7 $\pm$ 1.8&78.1 $\pm$ 2.3&94.8 $\pm$ 0.2&94.4 $\pm$ 0.2\\
			\bottomrule
		\end{tabular}
	}
	\caption{Results of ablation study on link prediction AUC (in \%), including S-VGAE using one and two GCN blocks, with and without (w/o.) skip-connections (Skip.). The best results are in bold.}\label{ablation}
\end{table}

\subsection{Ablation Study}

We also conduct ablation study to evaluate the effectiveness of different components in our proposed method. The link prediction experimental results are given in Table~\ref{ablation} (see the Appendix for more results). For comparisons between the S-VGAE method with one and two GCN blocks, the one-block version performs better on relatively small and dense graphs, such as Core and CiteSeer. While for larger and sparser graphs, such as PubMed and ogbl-collab, the two-block version with skip-connections has significantly improved model performances. The main reason is that in these large graphs, the one-hop low-order neighbor information are sparse and insufficient, where as our proposed two-block encoder with multiple spiking GNNs can enable the model to aggregate higher-order neighborhoods and improve the representation learning ability. We also demonstrate the superiority of decoupling the conventional GCN layers for energy efficiency in Table~\ref{compression} (see the Appendix for more results). The results show that the proposed spiking GCN blocks, including propagation and transformation layers, can greatly reduce the energy consumption by eliminating the MUL operations in the encoder.


\begin{table}[htbp]
	\centering
	\resizebox{0.9\columnwidth}{!}{
		\begin{tabular}{lcccc}
			\toprule
			\rule{0pt}{10pt}      &$\text{N}_{\AC}$&$\text{N}_{\MUL}$&$\text{E}^{\text{F}}$&$\text{E}^{\text{I}}$\\
			\midrule
			\rule{0pt}{10pt}S-VGAE&4.70&0.00&4.23         &0.47\\[2pt]
			\quad w/o. Decoupling&5.28&2.33&13.36        &7.74\\[2pt]
			Ratio                &--  &--  &3.15$\times$ &16.47$\times$\\
			\bottomrule
		\end{tabular}
	}
	\caption{Results of ablation study on number of MAC operations ($\times$10$^4$) and energy consumption ($\times$10$^4$pJ) to predict per link on the Cora dataset, including the proposed S-VGAE with and without the decoupling operation, and the improvement ratios of energy consumption.}
\label{compression}
\end{table}

\section{Conclusion}

In this paper, we propose the S-VGAE method for graph representation learning under the SNN framework. The proposed method consists of a spiking encoder with multiple GNN blocks and a probabilistic spiking decoder for the representative multi-node graph representation learning problem. To reduce the number of FLOPs and improve energy efficiency, we decouple conventional GNNs into the propagation and transformation operations. The spiking decoder leverages the W-IP readout layer for graph reconstruction.

\bibliography{ref}

\appendix

\section{Full Expression of the KL divergence}

We leverage the sigmoid surrogate gradient, thus the KL divergence is derived as
\begin{align}\label{kl}
	\KL[q(\z_n^{t}\vert\X^{1:t})\Vert p(\z_n^{t})]=&\sum_{c}^{C_z}z_{n,c}^{t}\log\frac{\sigma(u_{n,c}^t-V_{th})}{\pi}+\nonumber\\
	&(1-z_{n,c}^{t})\log\frac{1-\sigma(u_{n,c}^t-V_{th})}{1-\pi}.
\end{align}
The KL divergence serves as a regularization term to penalize the variational distributions that are far from the prior. Therefore, by setting a sparse Bernoulli prior with a small $\pi$, the decoder can generate sparse binary latent variables, which can improve the energy efficiency for SNNs. Also note that the KL term in Eq.~(13) is averaged over the time steps, so as to leverage the spatio-temporal backpropagation (STBP) \cite{wu2018spatio} algorithm for training.

\section{Algorithm}

The pseudo-code for training the proposed S-VGAE is given in Algorithm~\ref{algorithm}.

\begin{algorithm}[tb]
	\caption{Model Training for S-VGAE}
	\label{algorithm}
	\textbf{Input}: Undirected graph $\G$; node feature matrix $\X$\\
	\textbf{Parameter}: Learning rate $\kappa$; layer number $L$; time window $T$; other hyperparameters $\Theta$\\
	\textbf{Output}: Reconstructed graph $\hat{\G}$
	\begin{algorithmic}[1] 
		\STATE \textbf{Initialize} weight parameters $\W^l$, $l=1,\dots,L$, and $\w$
		\STATE $\HH^{1:T,0}=\textsc{Encoding}(\X)$
		\WHILE{training}
		\FOR{$t=1,\dots,T$}
		\FOR{$l=0,\dots,L-1$}
		\STATE $\tilde{\HH}^{t,l+1}=\textsc{Propagation}(\HH^{t,l},\Theta)$
		\STATE $\HH^{t,l+1}=\textsc{Transformation}(\tilde{\HH}^{t,l+1},\W^{l},\Theta)$
		\ENDFOR
		\STATE $\hat{\ppi}^{t}=\textsc{Linear}(\HH^{t,L},\W^{L})$
		\STATE $\Z^{t}=\textsc{BernoulliSampling}(\hat{\ppi}^{t},\Theta)$
		\STATE $\eeta^{t}=\textsc{WeightedInnerProduct}(\Z^{t},\w)$
		\ENDFOR
		\STATE $\hat{\G}=\textsc{Readout}(\eeta^{t},\Theta)$
		\FOR{$l=0,\dots,L$}
		\STATE $\W^l\gets\W^l-\kappa\nabla_{W^l}\mathcal{L}(\G,\hat{\G},\Z^{1:T},\Theta)$
		\ENDFOR
		\STATE $\w\gets\w-\kappa\nabla_{w}\mathcal{L}(\G,\hat{\G},\Z^{1:T},\Theta)$
		\ENDWHILE
	\end{algorithmic}
\end{algorithm}




\section{MAC Calculation}

In the experiments, we calculate the energy consumption of different methods as the product of the unit energy cost and the number of AC and MUL operations. Here we provide the detailed calculations of ACs and MULs for learning each node representation in different layers. Note that for link prediction, the operations of predicting each link should be twice as that of learning each node representation since we need to learn representations of both the source and target nodes.

\paragraph{Propagation}
The propagation layer of GCNs performs neighbor aggregation by multiplying a sparse normalized Laplacian matrix and the node feature matrix, thus the number of MUL operations for the $l$-th layer is
\begin{align}\label{ml_propagation}
	\text{N}_{\MUL}^{\PP_l}=TC_l(\vert\E\vert+N),
\end{align}
where $\vert\E\vert$ is the number of edges and $C_l$ is the dimension of feature channels. Typically, the number of AC operations can be obtained by $\text{N}_{\AC}^{\PP_l}=\text{N}_{\MUL}^{\PP_l}-1$ for ANNs. However, for SNNs, the features $\HH^{t,l}\in\R^{N\times C_l}$ are binary and usually sparse. Therefore, the number of AC operations for the spiking propagation layer can be further simplified as
\begin{align}\label{ac_propagation}
	\text{N}_{\AC}^{\PP_l}=\sum_{t}^{T}\frac{1}{N}\NN(\HH^{t,l})(\vert\E\vert+N),
\end{align}
where $\NN(\cdot)$ denotes the number of non-zero elements for a matrix. Intuitively, $	\text{N}_{\AC}^{\PP_l}$ can be significantly reduced when $\HH^{t,l}$ is very sparse, i.e., $\NN(\HH^{t,l})/N\ll C_l$.

\begin{table*}[!htbp]
	\centering
		\begin{tabular}{lc|cccc|cc|cc}
			\toprule
			\rule{0pt}{10pt}&&\multicolumn{4}{c|}{\underline{\textbf{ANN-Based}}}&\multicolumn{2}{c|}{\underline{\textbf{SNN-Based}}}&\multicolumn{2}{c}{\underline{\textbf{Our Methods}}}\\[2pt]
			&&GCN*&GAT*&VGAE&Graphite&GC-SNN*&SpikingGCN*&S-GAE&S-VGAE\\
			\midrule
			\rule{0pt}{10pt}\multirow{2}{*}{\textbf{Cora}}&$\text{E}^{\text{F}}$& 20.96$\times$ &62.66 $\times$&21.91$\times$&40.88$\times$&7.03$\times$&9.20$\times$&1.00$\times$&1.00$\times$\\[2pt]
			&$\text{E}^{\text{I}}$&130.45$\times$&390.07$\times$&136.37$\times$&253.56$\times$&42.53$\times$&54.87$\times$&\underline{1.06$\times$}&1.00$\times$\\
			\midrule
			\rule{0pt}{10pt}\multirow{2}{*}{\textbf{CiteSeer}}&$\text{E}^{\text{F}}$& 38.33 $\times$ &76.60$\times$&39.01$\times$&55.19$\times$&6.36$\times$&7.72$\times$&1.00$\times$&1.00$\times$\\[2pt]
			&$\text{E}^{\text{I}}$&238.97$\times$&477.59$\times$&243.24$\times$&343.77$\times$&38.92$\times$&44.98$\times$&1.09$\times$&1.00$\times$\\
			\midrule
			\rule{0pt}{10pt}\multirow{2}{*}{\textbf{PubMed}}&$\text{E}^{\text{F}}$&4.39$\times$&13.04$\times$&4.92$\times$&43.82$\times$&6.49$\times$&2.78$\times$&1.00$\times$&1.00$\times$\\[2pt]
			&$\text{E}^{\text{I}}$&27.42$\times$&81.40$\times$&30.70$\times$&272.97$\times$&39.75$\times$&17.21$\times$&1.00$\times$&1.00$\times$\\
			\midrule
			\rule{0pt}{10pt}\multirow{2}{*}{\textbf{ogbl-collab}}&$\text{E}^{\text{F}}$&3.10$\times$&77.72$\times$&4.65$\times$&82.14$\times$&3.75$\times$&1.59$\times$&1.00$\times$&1.00$\times$\\[2pt]
			&$\text{E}^{\text{I}}$&19.36$\times$&485.47$\times$&29.02$\times$&511.63$\times$&20.87$\times$&8.92$\times$&1.00$\times$&1.00$\times$\\
			\bottomrule
		\end{tabular}
	\caption{The relative energy cost of floating-point ($\text{E}^{\text{F}}$) and integer ($\text{E}^{\text{I}}$) MAC operations for different methods. The results of our proposed S-VGAE are set to be one, and those of the comparative methods are multiples of S-VGAE. The asterisk * indicates that the method has been modified using the GAE framework.}\label{ratio}
\end{table*}

\paragraph{Transformation}
The transformation layer performs linear transformation to the node feature matrix $\HH^{t,l}$ using trainable weights $\W^l\in\R^{C_{l}\times C_{l+1}}$. The number of MUL operations is given as
\begin{align}\label{ml_transformation}
	\text{N}_{\MUL}^{\T_l}=TNC_lC_{l+1}.
\end{align}
Similarly, the number of AC operations can be calculated as $\AC^l_{T}=\MUL^l_{T}-1$. When the feature matrix is sparse, it can be simplified as
\begin{align}\label{ac_transformation}
	\text{N}_{\AC}^{\T_l}=\sum_{t}^{T}\frac{1}{N}\NN(\HH^{t,l})C_{l+1}.
\end{align}
For conventional GCN aggregators, the number of MUL operations can be obtained by $\text{N}_{\MUL}^{\GC_l}=\text{N}_{\AC}^{\PP_l}+\text{N}_{\MUL}^{\T_l}$. However, since the outputs of propagation are continuous signals, the MUL operations cannot be transformed to ACs, and $\AC^l_{T}$ can no longer be reduced using Eq.~\ref{ac_transformation}, either. Therefore, the decoupling of propagation and transformation can significantly reduce FLOPs and improve energy efficiency.

\paragraph{Weighted Inner Product} The number of AC operations for the proposed weighted inner product layer predicting each link is
\begin{align}\label{ac_inner}
	\text{N}_{\AC}^{\WIP}=\sum_{t}^{T}\frac{1}{N}\NN^2(\Z^{t}).
\end{align}
And the number of MUL operations for the weighted inner product layer of SNNs is $\text{N}_{\MUL}^{\WIP}=0$.

\section{More Details of Experimental Results}

In this section, we provide more results of the experiments, including the number of FLOPs for different methods to predict a single link, more results of the ablation study, and a sensitivity analysis for some important hyperparameters of our method.

\subsection{Energy Efficiency Analysis}

\begin{figure*}[!htbp]
	\centering
	\subfloat[Cora]{
		\includegraphics[width=0.24\textwidth]{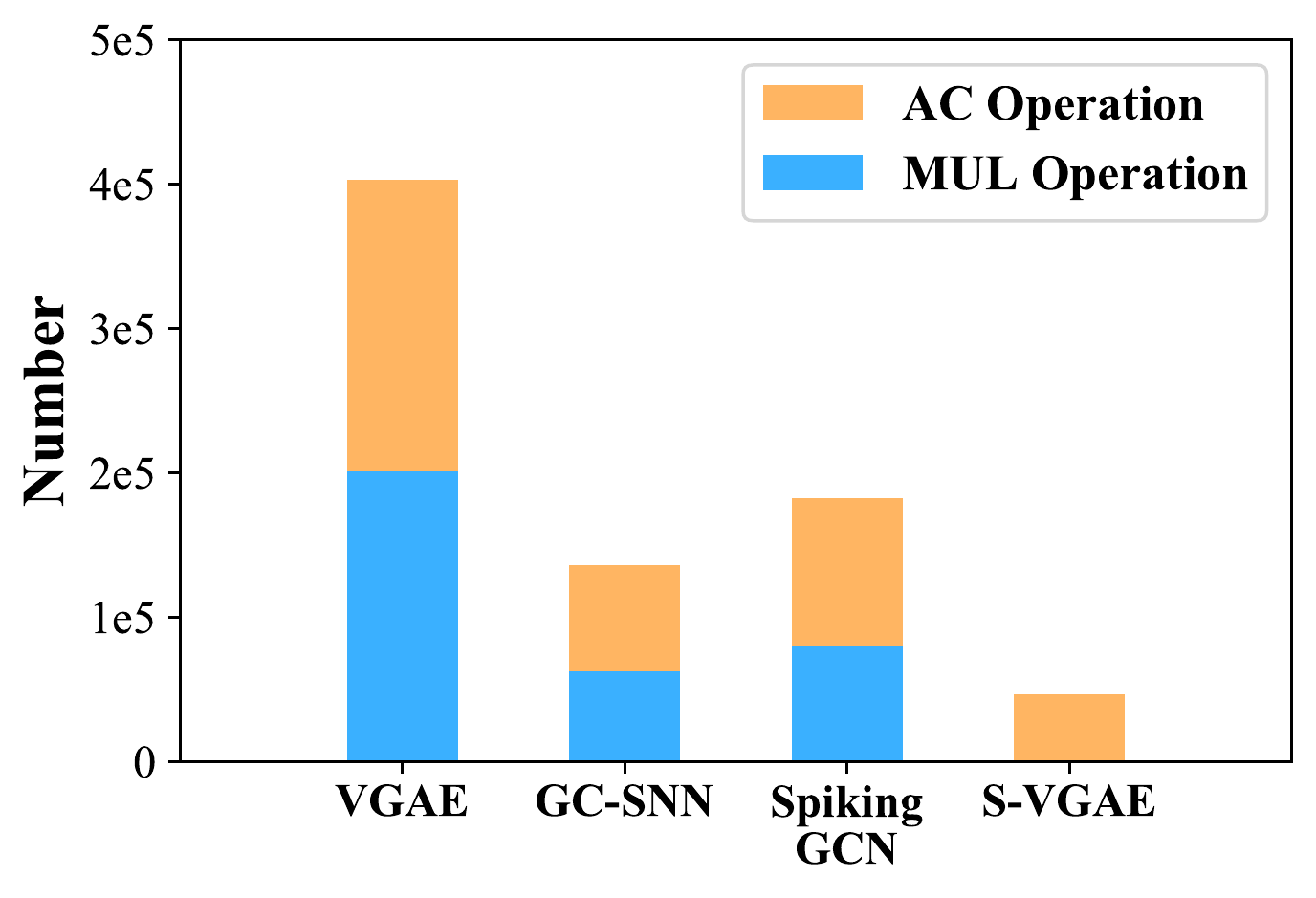}
	}
	\subfloat[CiteSeer]{
		\includegraphics[width=0.24\textwidth]{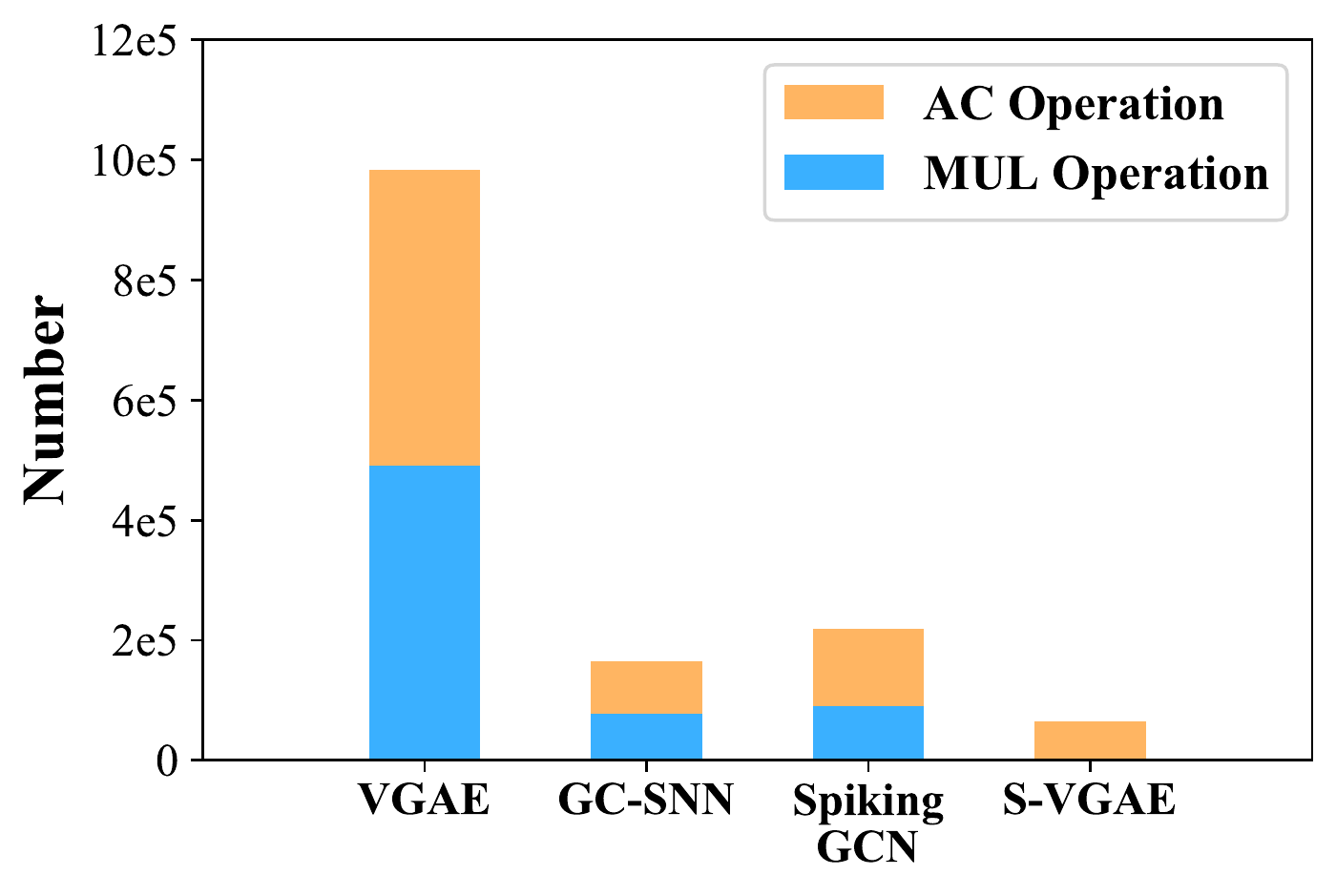}
	}
	\subfloat[PubMed]{
		\includegraphics[width=0.24\textwidth]{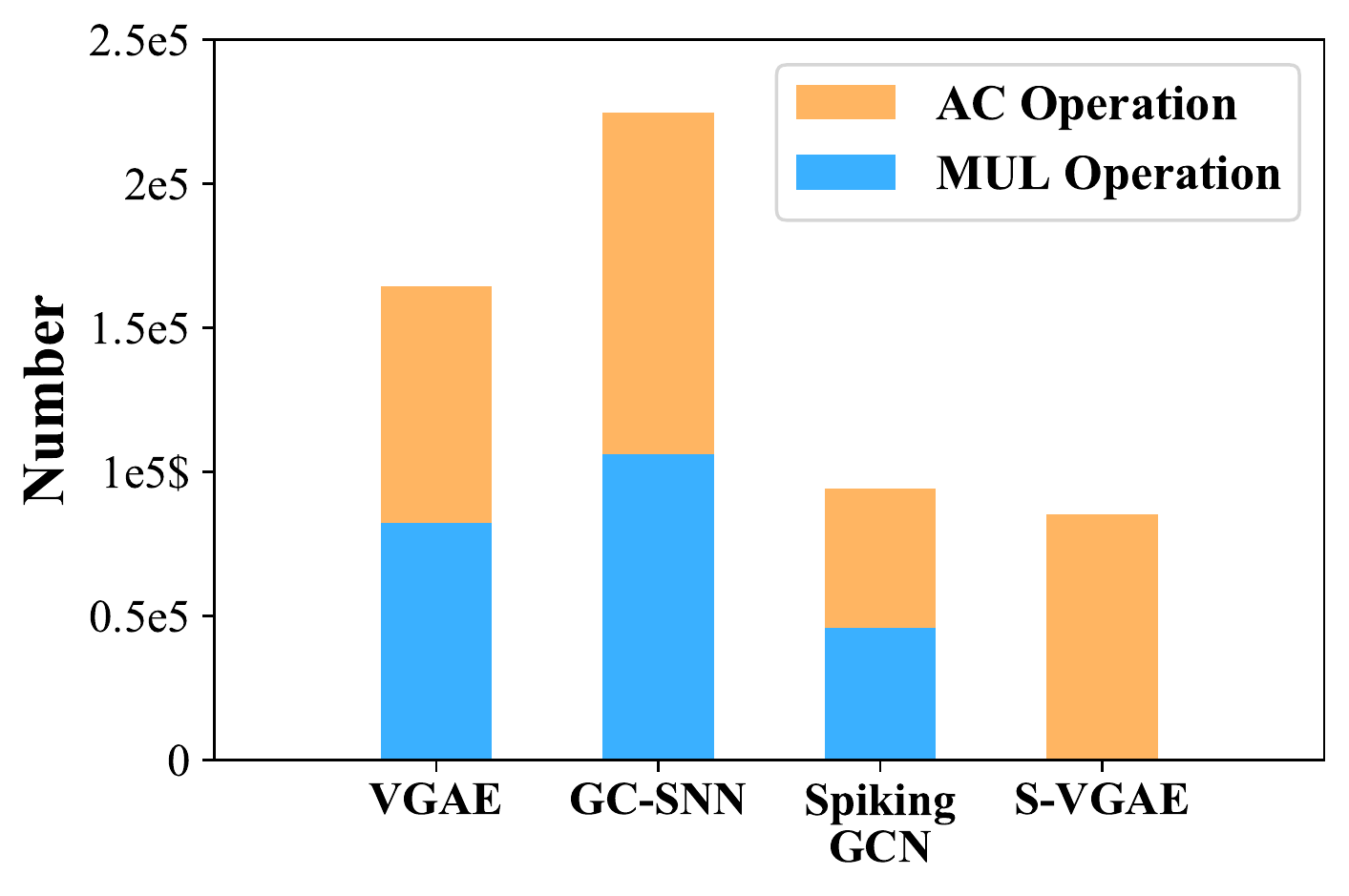}
	}
	\subfloat[ogbl-collab]{
		\includegraphics[width=0.24\textwidth]{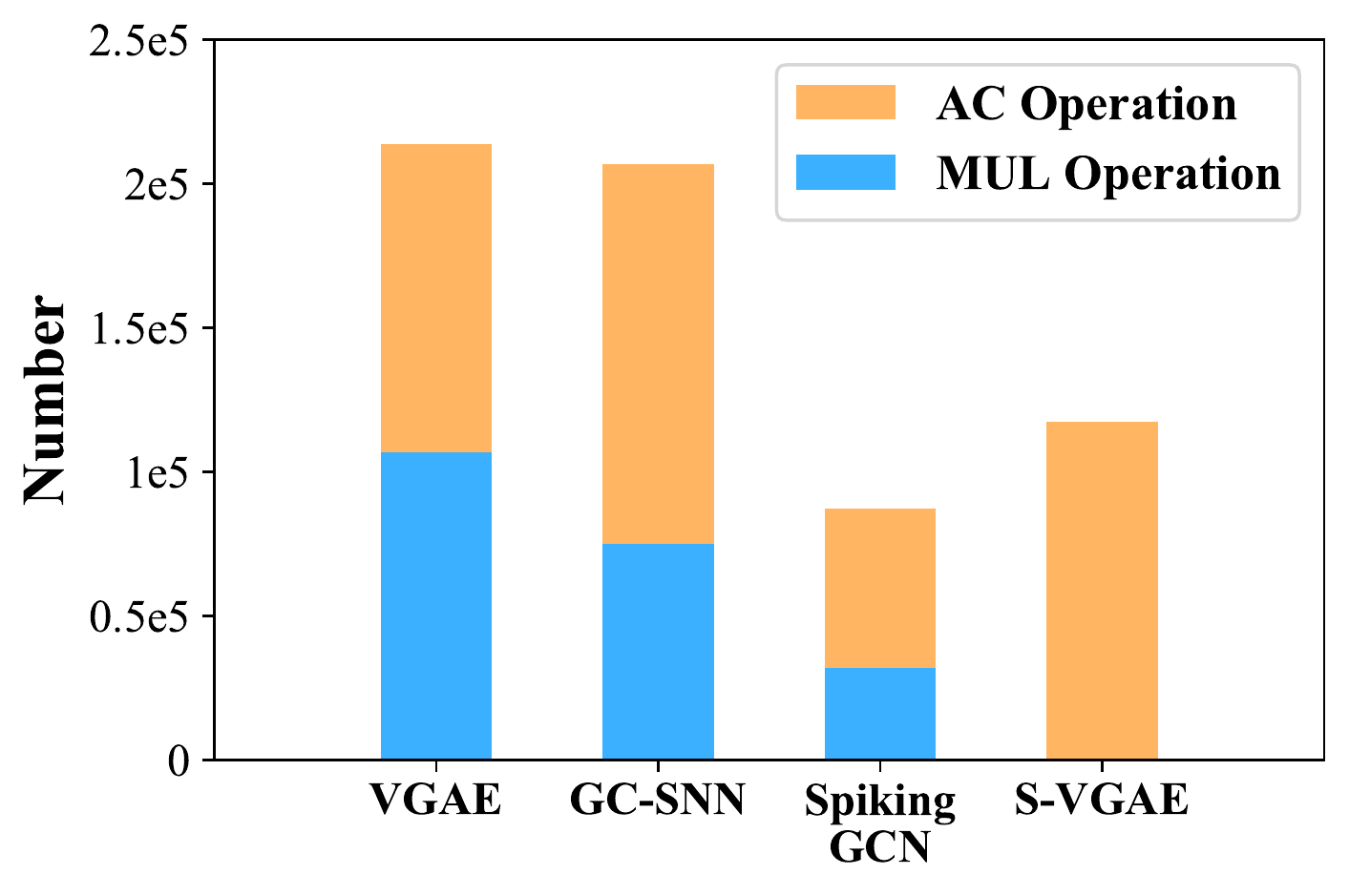}
	}
	\centering
	\caption{The number of FLOPs for different methods to predict a single link on average. The orange and blue colors stand for accumulate (AC) and multiply (MUL) operations, respectively.}\label{flops}
\end{figure*}

The experimental results for the average FLOPs of different methods are illustrated in Fig.~\ref{flops}. The comparative methods, including the ANN-based VGAE as well as the SNN-based GC-SNN and SpikingGCN, require similar numbers of AC and MUL operations, whereas the energy consumption of the latter accounts for a vast majority of the total consumption (see Fig.~3 in the main text). Our proposed S-VGAE, in contrast, only has the more energy-saving AC operations and thus is much more efficient. As a more intuitive comparison, we further present the relative energy cost between the comparative methods and S-VGAE in Table~\ref{ratio}. The results show that our proposed methods without any expensive MUL operation can significantly compress the energy consumption of link prediction compared with both ANN- and SNN-based methods.

\begin{figure*}[!htbp]
	\begin{center}
		\subfloat[$T$]{
			\includegraphics[width=0.24\textwidth]{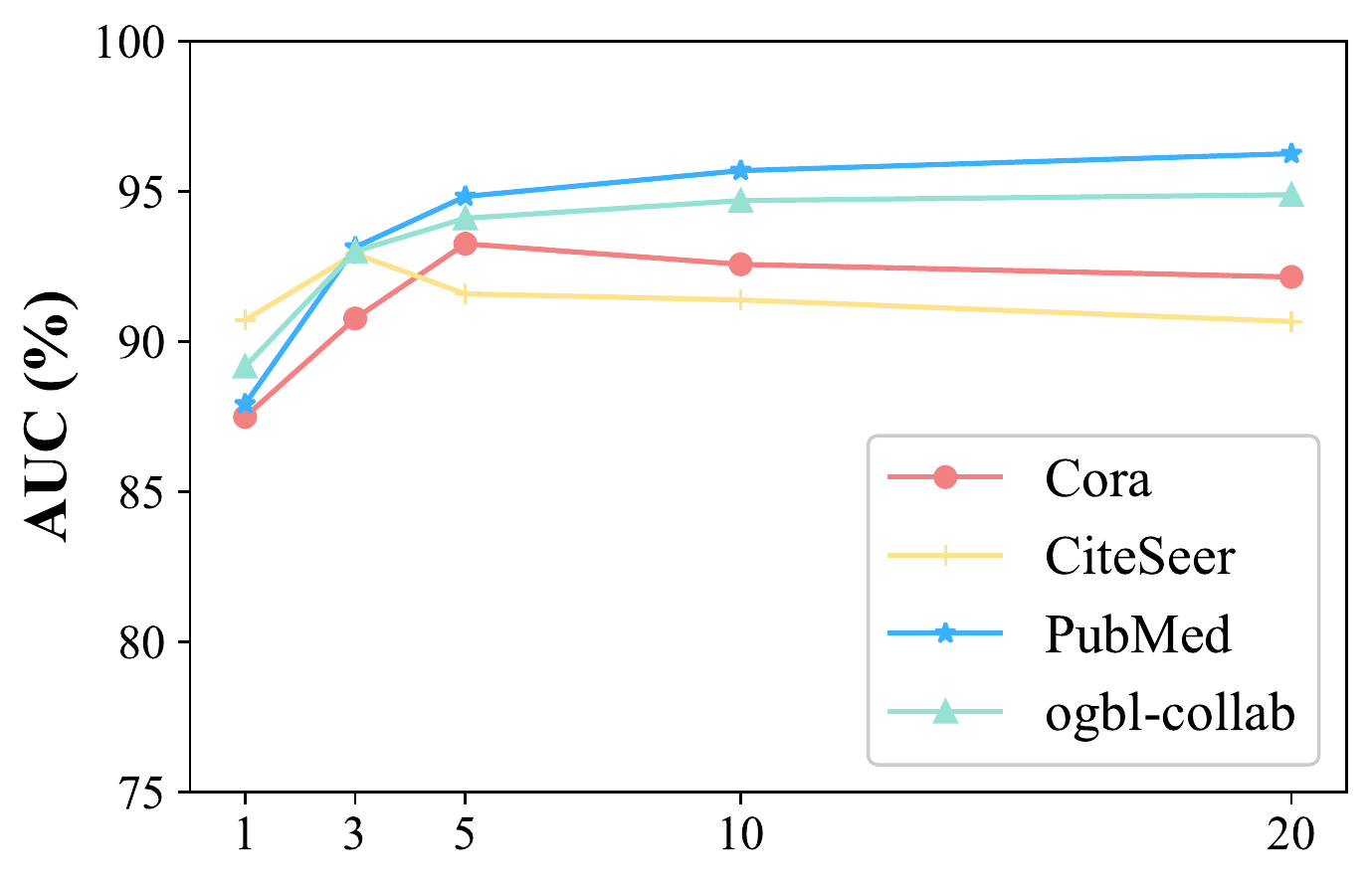}
		}
		\subfloat[$T$]{
			\includegraphics[width=0.24\textwidth]{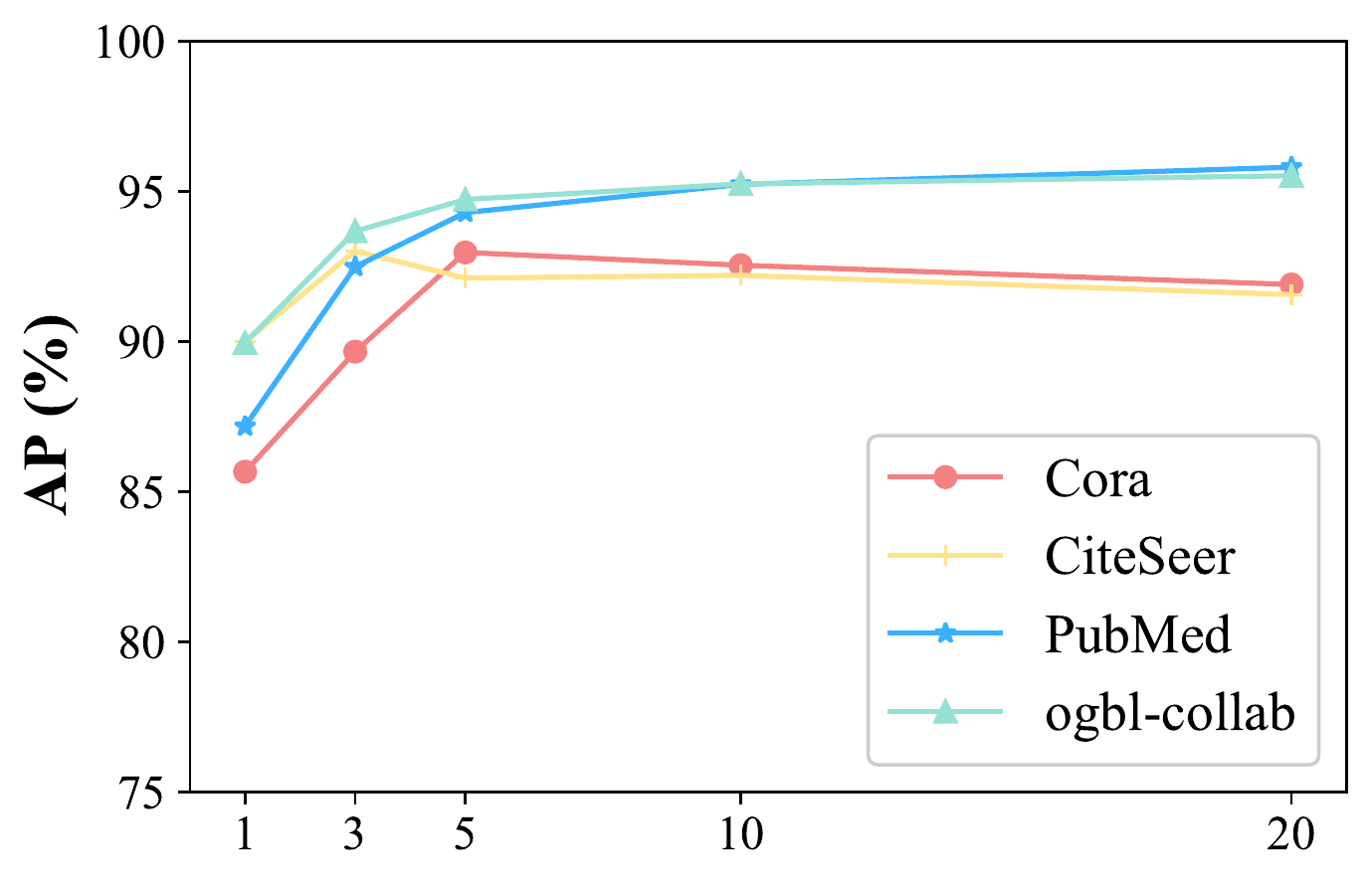}
		}
		\subfloat[$V_{th}$]{
			\includegraphics[width=0.24\textwidth]{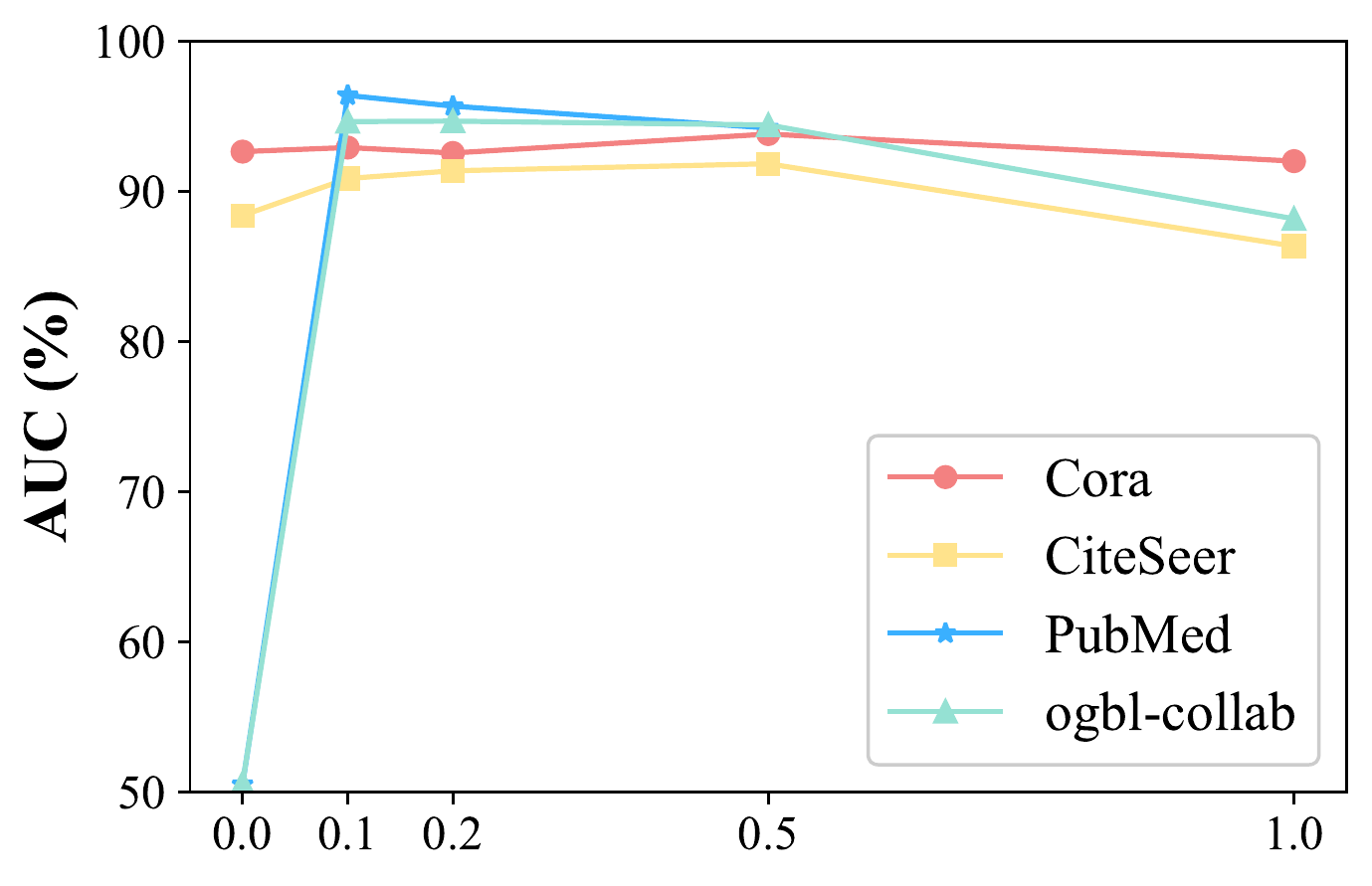}
		}
		\subfloat[$V_{th}$]{
			\includegraphics[width=0.24\textwidth]{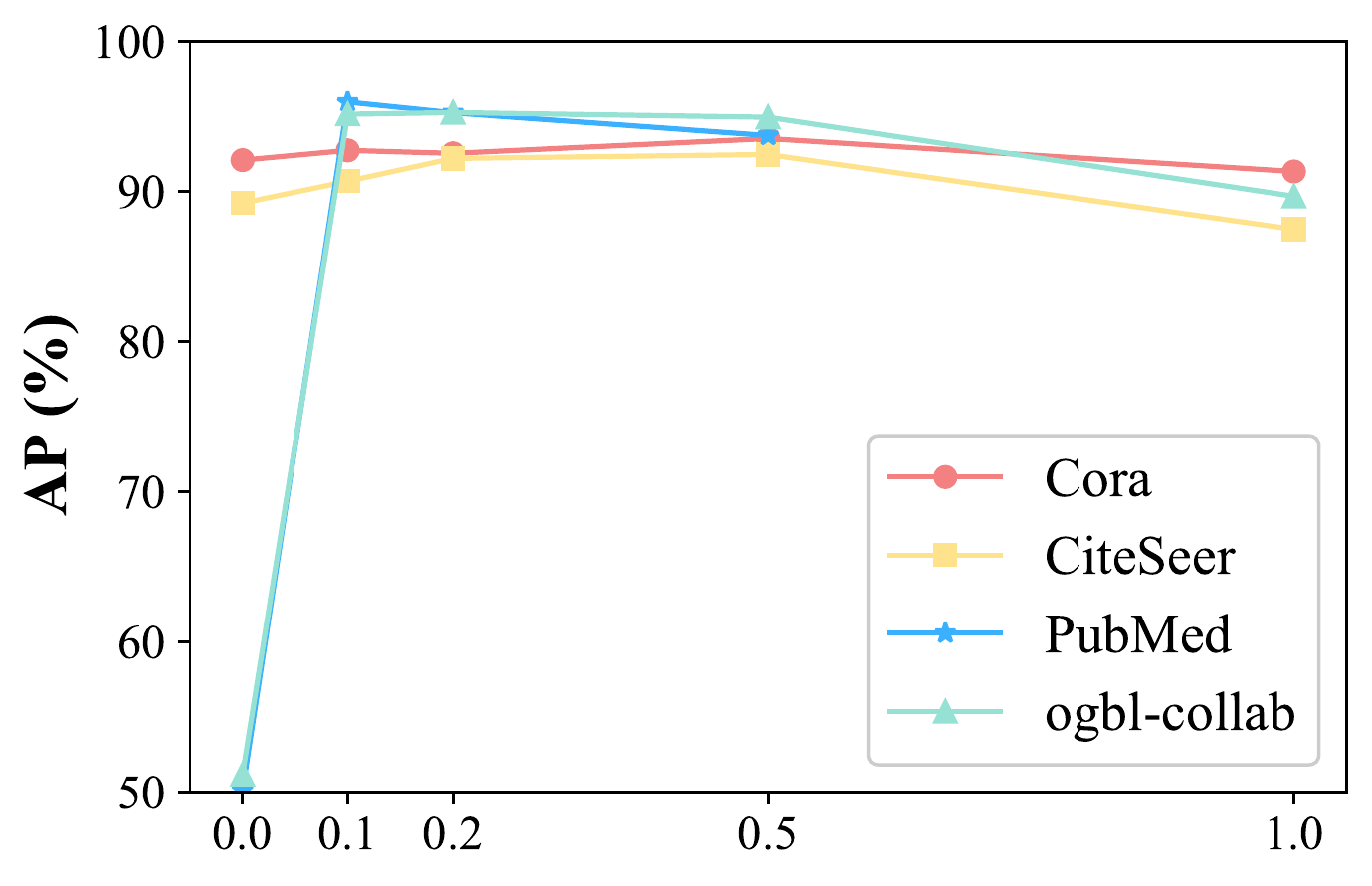}
		}
		\centering
		\caption{Results of sensitivity analysis for the time window $T$ (a and b) and firing threshold $V_{th}$ (c and d).}\label{sensitivity}
	\end{center}
\end{figure*}

\subsection{Ablation Study}

Here we present more experimental results of the ablation study. Table~\ref{ablation2} shows the AP scores of link prediction. Table~\ref{compression2} shows the number of MAC operations and the total energy consumption of floating points and integers on different datasets, including CiteSeer, PubMed and ogbl-collab.

\begin{table}[htbp]
	\centering
	\resizebox{\columnwidth}{!}{
		\begin{tabular}{lcccc}
			\toprule
			\rule{0pt}{10pt}         &\textbf{Cora}          &\textbf{CiteSeer}      &\textbf{PubMed}&\textbf{ogbl-collab}\\
			\midrule
			\rule{0pt}{10pt}One-Block&\textbf{92.6 $\pm$ 0.5}&\textbf{92.1 $\pm$ 0.9}&93.3 $\pm$ 0.2&89.8 $\pm$ 1.0\\[2pt]
			Two-Block                &90.8 $\pm$ 2.3   &84.5 $\pm$ 0.5&\textbf{95.6 $\pm$ 0.2}&\textbf{95.6 $\pm$ 0.5}\\[2pt]
			\quad w/o. Skip.&81.6 $\pm$ 1.9&78.3 $\pm$ 2.9&95.2 $\pm$ 0.2&95.0 $\pm$ 0.3\\
			\bottomrule
		\end{tabular}
	}
	\caption{Results of ablation study on link prediction AP (in \%), including S-VGAE using one and two GCN blocks, with and without (w/o.) skip-connections (Skip.). The best results are in bold.}\label{ablation2}
\end{table}

\begin{table}[!htbp]
	\centering
	\resizebox{0.9\columnwidth}{!}{
		\begin{tabular}{lcccc}
			\toprule
			\rule{0pt}{10pt}      &\multicolumn{4}{c}{\textbf{CiteSeer}}\\
			\midrule
			\rule{0pt}{10pt}      &$\text{N}_{\AC}$&$\text{N}_{\MUL}$&$\text{E}^{\text{F}}$&$\text{E}^{\text{I}}$\\
			\midrule
			\rule{0pt}{10pt}S-VGAE&6.44            &0.00             &5.80                 &0.65\\[2pt]
			\quad w/o. Decoupling &2.03            &1.21             &6.32                 &3.97\\[2pt]
			Ratio                 &--              &--               &1.09$\times$         &6.11$\times$\\
			\midrule\midrule
			\rule{0pt}{10pt}      &\multicolumn{4}{c}{\textbf{PubMed}}\\
			\midrule
			\rule{0pt}{10pt}      &$\text{N}_{\AC}$&$\text{N}_{\MUL}$&$\text{E}^{\text{F}}$&$\text{E}^{\text{I}}$\\
			\midrule
			\rule{0pt}{10pt}S-VGAE&8.55            &0.00             &7.70                 &0.86\\[2pt]
			\quad w/o. Decoupling &13.15           &7.32             &38.92                &24.01\\[2pt]
			Ratio                 &--              &--               &5.05$\times$         &27.92$\times$\\
			\midrule\midrule
			\rule{0pt}{10pt}      &\multicolumn{4}{c}{\textbf{ogbl-collab}}\\
			\midrule
			\rule{0pt}{10pt}      &$\text{N}_{\AC}$&$\text{N}_{\MUL}$&$\text{E}^{\text{F}}$&$\text{E}^{\text{I}}$\\
			\midrule
			\rule{0pt}{10pt}S-VGAE&11.75           &0.00             &10.58                &1.18\\[2pt]
			\quad w/o. Decoupling &21.83           &4.79             &37.37                &17.03\\[2pt]
			Ratio                 &--              &--               &3.53$\times$         &14.43$\times$\\
			\bottomrule
		\end{tabular}
	}
	\caption{Results of ablation study on the number of MAC operations ($\times$10$^4$) and energy consumption ($\times$10$^4$pJ) to predict one single link on different datasets, including the proposed S-VGAE with and without the decoupling operation, and the improvement ratios of energy consumption.}
	\label{compression2}
\end{table}

\subsection{Sensitivity Analysis}

To evaluate the robustness of S-VGAE on important hyperparameters, we conduct sensitivity analysis for the time window $T$ and firing threshold $V_{th}$ on different datasets. The experimental results are presented in Fig.~\ref{sensitivity}.

Specifically, Fig.~\ref{sensitivity}(a) and (b) demonstrate that, based on the directly trained SNN framework with surrogate gradient, our proposed S-VGAE method achieves low latency with less than 10 time steps. Furthermore, from Fig.~\ref{sensitivity}(c) and (d), it can be seen that our S-VGAE enjoys good robustness to the threshold parameter on Cora and CiteSeer due to the proposed probabilistic LIF model, which takes the membrane potential as the log-odds for a neuron to emit spikes and thus turn the firing threshold to be soft. While on the large graphs such as PubMed and ogbl-collab, our method is more sensitive to a small firing threshold (e.g., $V_{th}=$0). One possible reason is that these large graphs can aggregate more node features during propagation, which leads to high membrane potential accumulated in spiking neurons. As a result, a small threshold may account for too many firings and reduce the variability of features.

\end{document}